\documentclass[pdflatex,sn-mathphys-num]{sn-jnl}

\usepackage{graphicx} % Required for inserting images
\usepackage{booktabs}
\usepackage{tabularx}
\usepackage{makecell}
\usepackage{bbding}
\usepackage{romannum}
\usepackage{pifont}
\usepackage{color,colortbl}
\usepackage{amsmath}
\usepackage{amssymb}
\usepackage{hyperref}
\usepackage{layouts}
\usepackage{microtype}

\usepackage[table,xcdraw]{xcolor}

\usepackage{pdflscape}   % 横向页面
\usepackage{booktabs}
\usepackage{multirow}
\usepackage{array}
\usepackage{amssymb}     % \checkmark \times
\usepackage{amsmath}
\usepackage{geometry}
\begin{document}

\title[Article Title]{Building Egocentric Procedural AI Assistant: Methods, Benchmarks, and Challenges \vspace{1em}}

%\text{\\}

\author[1]{\sur{Junlong Li}}

\author[1]{\sur{Huaiyuan Xu}}

\author[2]{\sur{Sijie Cheng}}

\author[3]{\sur{Kejun Wu}}

\author[4]{\sur{Kim-Hui Yap}}

\author[1]{\sur{Lap-Pui Chau}}

\author*[1]{\sur{Yi Wang}}\email{yi-eie.wang@polyu.edu.hk (wang1241@e.ntu.edu.sg)}

%%\affil[1]{\orgdiv{Department of Electrical and Electronic Engineering}, \orgname{The Hong Kong Polytechnic University}, \orgaddress{\city{Hong Kong SAR}}}
\affil[1]{\orgdiv{Department of EEE}, \orgname{The Hong Kong Polytechnic University}, \orgaddress{\city{Hong Kong SAR}}}

\affil[2]{\orgname{RayNeo}, \orgaddress{\city{Shenzhen}}; \orgname{Tsinghua University}, \orgaddress{\city{Beijing}, \country{China}}}

%%\affil[3]{\orgdiv{School of Electronic Information and Communications}, \orgname{Huazhong University of Science and Technology}, \orgaddress{\city{Wuhan}, \country{China}}}
\affil[3]{\orgdiv{School of EIC}, \orgname{Huazhong University of Science and Technology}, \orgaddress{\city{Wuhan}, \country{China}}}

%%\affil[4]{\orgdiv{School of Electrical and Electronic Engineering}, \orgname{Nanyang Technological University}, \orgaddress{ \country{Singapore}}}
\affil[4]{\orgdiv{School of EEE}, \orgname{Nanyang Technological University}, \orgaddress{ \country{Singapore}}}

\date{November 2025}

%\setcounter{page}{0}
%\raggedbottom
\pagenumbering{arabic}

%\begin{center} {\LARGE \textbf{Building Egocentric Procedural AI Assistant: Methods, Benchmarks, and Challenges}}\\[1em] {\large Junlong Li\textsuperscript{1}, Huaiyuan Xu\textsuperscript{1}, Sijie Cheng\textsuperscript{2}, Kejun Wu\textsuperscript{3}, Kim-Hui Yap\textsuperscript{4}, Lap-Pui Chau\textsuperscript{1}, Yi Wang\textsuperscript{1,*}\\[0.5em] } {\small \textsuperscript{1}Department of EEE, The Hong Kong Polytechnic University, Hong Kong SAR\\ \textsuperscript{2}RayNeo, Shenzhen; Tsinghua University, Beijing, China\\ \textsuperscript{3}School of EIC, Huazhong University of Science and Technology, Wuhan, China\\ \textsuperscript{4}School of EEE, Nanyang Technological University, Singapore\\ \textsuperscript{*}Email: yi-eie.wang@polyu.edu.hk }[1em] \end{center} 

\abstract{Driven by recent advances in vision-language models (VLMs) and egocentric perception research, the emerging topic of an egocentric procedural AI assistant (EgoProceAssist) is introduced to step-by-step support daily procedural tasks in a first-person view. In this paper, we start by identifying three core tasks in EgoProceAssist: egocentric procedural error detection, egocentric procedural learning, and egocentric procedural question answering, then introduce two enabling dimensions: real-time and streaming video understanding, and proactive interaction in procedural contexts. We define these tasks within a new taxonomy as the EgoProceAssist's essential functions and illustrate how they can be deployed in real-world scenarios for daily activity assistants. Specifically, our work encompasses a comprehensive review of current techniques, relevant datasets, and evaluation metrics across these five core areas. To clarify the gap between the proposed EgoProceAssist and existing VLM-based assistants, we conduct novel experiments to provide a comprehensive evaluation of representative VLM-based methods. Through these findings and our technical analysis, we discuss the challenges ahead and suggest future research directions. Furthermore, an exhaustive list of this study is publicly available in an active repository that continuously collects the latest work: \url{https://github.com/z1oong/Building-Egocentric-Procedural-AI-Assistant}.
}


\maketitle

\section{Introduction}

Procedural tasks, which require adherence to specific sequences of steps, are central to numerous daily and industrial activities. Errors in the order or execution of these steps can lead to inefficiencies, suboptimal outcomes, or even safety risks. Traditional video analysis methods, while capable of visual pattern recognition, lack a deep understanding of video content. They struggle to comprehend logical procedural narratives and face challenges in integrating multifunctional assistance, such as detection and video question answering (VQA). Recent advances in  VLM \cite{cheng2024videollama}\cite{lin2024video}\cite{lillava} and egocentric (first-person view) vision technologies \cite{thatipelli2025egocentric}\cite{li2025challenges}, combined with augmented and virtual reality devices, have enabled the development of egocentric AI assistants for procedural task guidance \cite{yang2025egolife,huang2025vinci}. 

AI assistants with egocentric vision acquire environmental observations through head-mounted cameras and AI glasses \cite{plizzari2024outlook}. Imagine an AI assistant capable of autonomously learning key procedural steps, detecting errors in real time, and providing contextual responses to user queries. Such a system holds significant potential across various fields, including industrial assembly \cite{ding2023every}, smart homes \cite{plizzari2024outlook}, remote medical assistance, and vocational training. 

Based on the envisioned functions, such as the ability to detect procedural errors in real time, the ability to learn the key steps of procedural tasks independently from videos, the ability to understand egocentric videos and answer relevant questions, this paper identifies three new technical approaches and two enabling dimensions, to address key requirements for building an egocentric procedural AI assistant (EgoProceAssist): egocentric procedural error detection, egocentric procedural learning, egocentric procedural question answering, real-time and streaming video understanding, and proactive interaction in procedural contexts. The main contributions of this work can be summarized as follows:

\begin{figure}[t]
    \centering
    \includegraphics[width=1\linewidth]{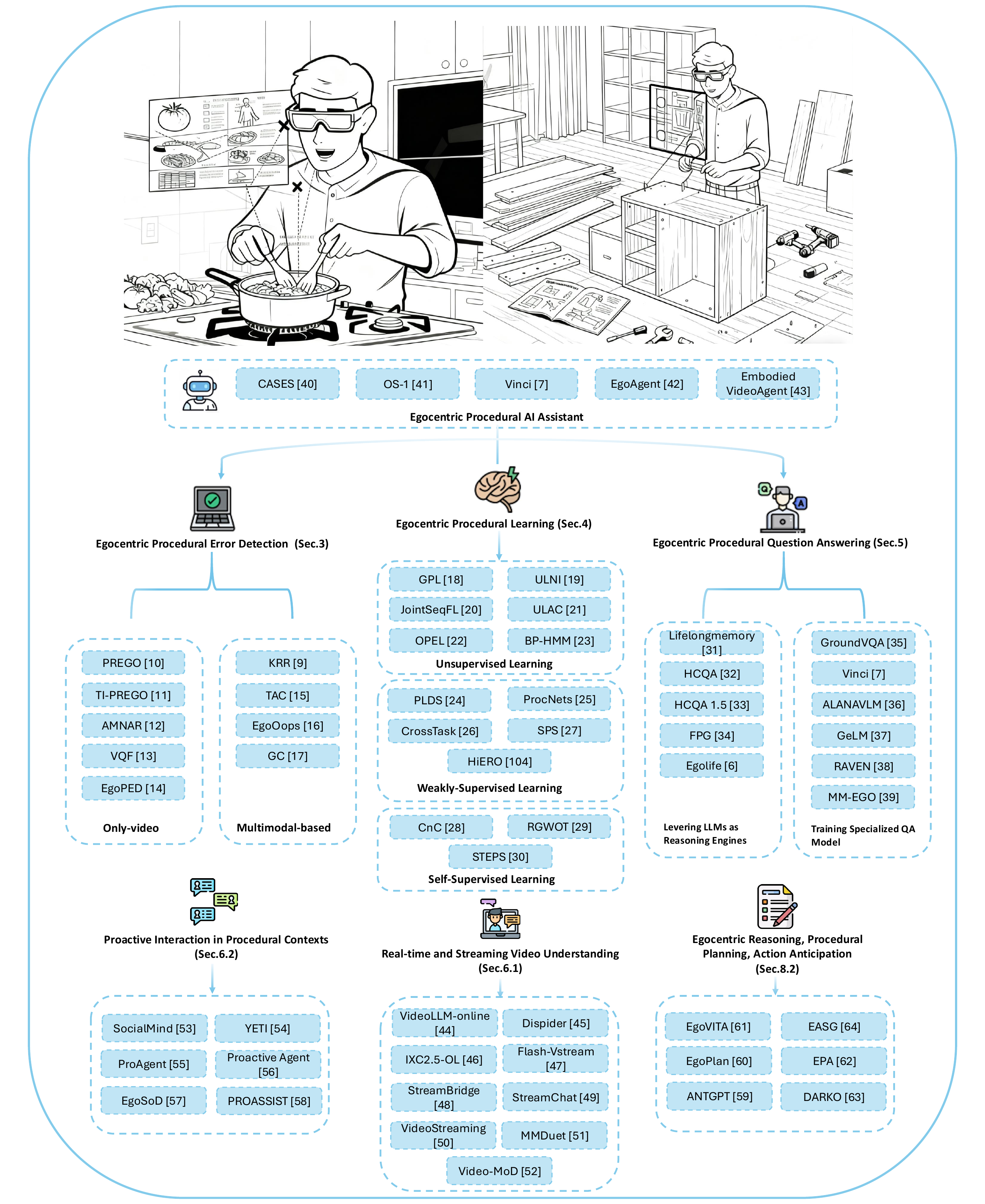}
    \caption{Taxonomy of EgoProceAssist \cite{flaborea2024prego,plini2025ti,huang2025modeling,patsch2025technical,lee2024error}\cite{ding2023every}\cite{storks2024transparent,haneji2025egooops,mazzamuto2025gazing,bansal2024united,alayrac2016unsupervised,elhamifar2019unsupervised,kukleva2019unsupervised,chowdhury2024opel,sener2015unsupervised,lin2022learning,zhou2018towards,zhukov2019cross,naing2020procedure,bansal2022my,mahmood2025procedure,shah2023steps,wang2023lifelongmemory,zhang2024hcqa,zhang2025hcqa,taluzzi2025pixels}\cite{yang2025egolife}\cite{di2024grounded}\cite{huang2025vinci}\cite{suglia2024alanavlm,chen2025grounded,biswas2025raven,yemmego}\cite{10.1145/3610910,10.1145/3659600,chen2025egoagentjointpredictiveagent,fan2025embodiedvideoagentpersistentmemory,chen2024videollm,qian2025dispider,zhang2024internlm,zhang2024flash,wang2025streambridge,xiongstreaming,qian2024streaming,wang2024videollm,wu2024videollm,yang2025socialmind,bandyopadhyay2025yeti,zhang2024proagent,luproactive,wang2025egosocial,zhang2025proactive,zhao2023antgpt,fang2024egocentric,kulkarni2025egovita,liu2023egocentric,rhinehart2017first,rodin2024action}.}
    \label{fig1}%
\end{figure}

\begin{figure*}[t]
    \centering
    \includegraphics[width=1\linewidth]{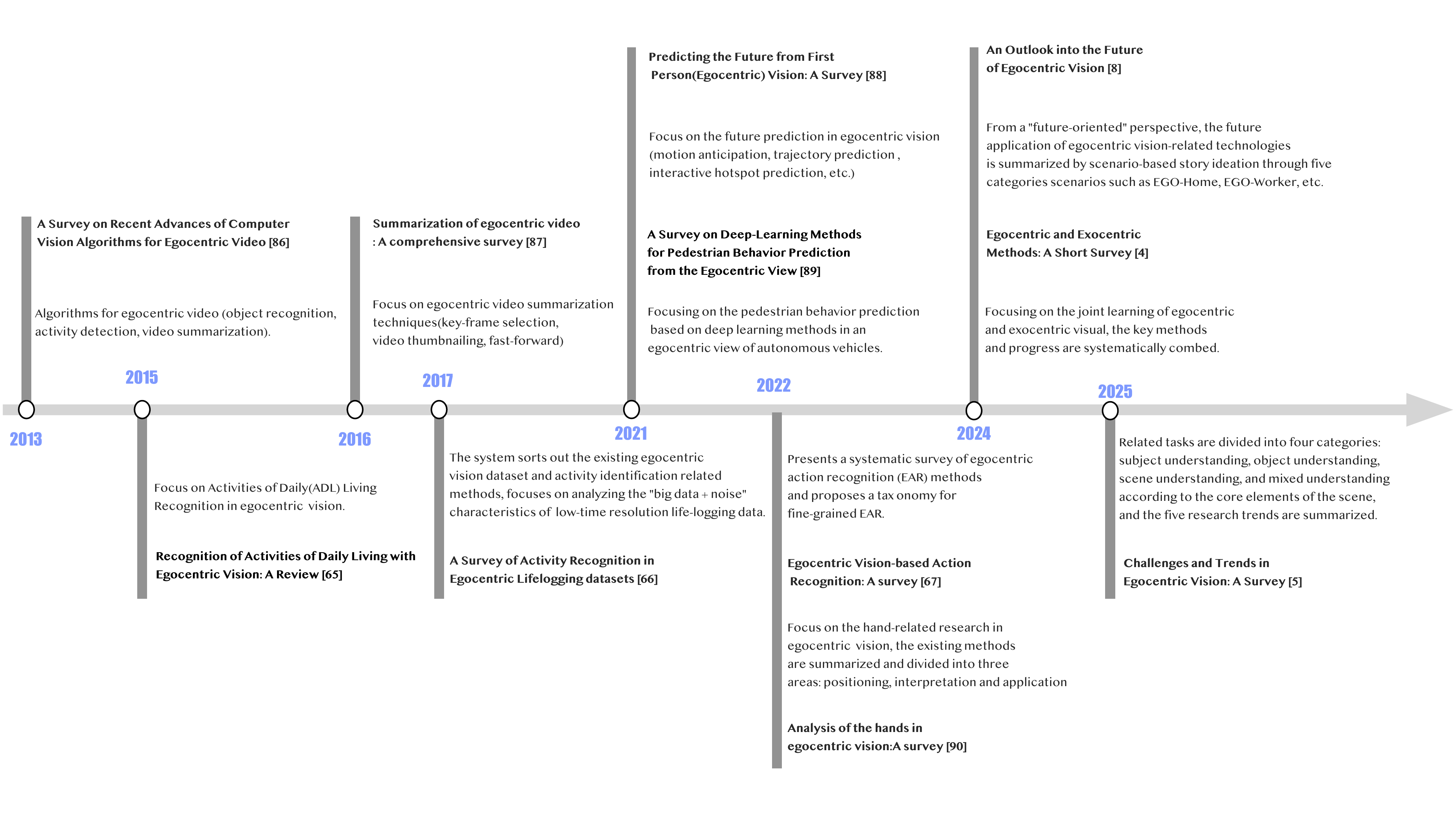}

        \caption{A timeline with the surveys in egocentric vision \cite{bambach2015survey}\cite{nguyen2016recognition}\cite{del2016summarization}\cite{hamid2017survey}\cite{rodin2021predicting}\cite{chen2021survey}\cite{nunez2022egocentric}\cite{bandini2020analysis}\cite{plizzari2024outlook}\cite{thatipelli2025egocentric}\cite{li2025challenges}.}
    \label{fig3}%
    \vspace{-3mm}
\end{figure*}

$\bullet$ Most previous research efforts concentrated on video understanding tasks from a third-person perspective, and there are relatively few surveys focused on video analysis in egocentric vision, as shown in Fig.~\ref{fig3}. Some of them focus on the field's action recognition technology \cite{nguyen2016recognition,hamid2017survey,nunez2022egocentric}, challenges \cite{li2025challenges}, and prospects \cite{plizzari2024outlook}. Therefore, a notable absence remains in the systematic survey literature addressing the development of AI assistants for procedural tasks from an egocentric perspective, particularly in the three critical domains of egocentric procedural error detection, egocentric procedural learning, and egocentric procedural question answering. This paper presents the first comprehensive survey of the state-of-the-art techniques in these areas.

$\bullet$ We present novel method taxonomies across the five domains, as shown in Fig.~\ref{fig1}, enabling systematic classification and synthesis with improved clarity and a unified understanding. Our survey analyzes 39 commonly used datasets and evaluation metrics relevant to these areas.

$\bullet$ To provide objective evidence and highlight the development potential of this field, we conduct two supplementary experiments evaluating recent VQA models and AI assistants across four datasets. The comparative analysis reveals key limitations of current AI assistants. 

$\bullet$ Finally, we discuss significant challenges and propose directions for future research, emphasizing the essential contribution of these domains to AI assistant development. Our synthesis aims to inspire future work in the field.

The remainder of this paper is organized as follows. In Section 2, domains and techniques related to the construction of EgoProceAssist are presented. In Sections 3, 4, 5, and 6, we provide a comprehensive summary of the existing technical approaches, commonly used datasets, and evaluation metrics for the three core tasks and two enabling dimensions, respectively. For enhanced clarity, we also present comparative tables to highlight performance differences among the methods. Section 7 presents experimental investigations assessing the capability of existing models to understand procedural tasks across two distinct domains. Section 8 discusses the current challenges faced in the field and explores potential trends for future research and development. Finally, Section 9 offers a comprehensive summary of the findings and conclusions drawn from this work.

\section{Primer On Egocentric Procedural AI Assistant }

This section consists of two parts. We first review the current progress in related fields, then structure EgoProceAssist around three function-oriented domains and two enabling dimensions.

\subsection{Related Work}

Progressing research in these areas will play a key role in developing EgoProceAssist, as they offer rich opportunities for further innovation. Methods in these domains leverage proven computer vision and deep learning techniques, enabling robust, scalable improvement and deployment.

\textit{\textbf{AI Assistants on Wearable Devices.}} As wearable devices become more widely deployed, research on AI assistants built on these platforms has grown. StretchAR \cite{10.1145/3550305} introduces tactile feedback via smart wristbands, and PrISM-QA \cite{10.1145/3699759} uses smartwatches for voice-based conversational assistance. Although these systems effectively exploit wearable devices, they make only limited use of visual input and thus are less suitable for tasks requiring rich real-world perception. LiSee \cite{10.1145/3550282} employs a neck-worn earphone to support visually impaired users in daily object retrieval using combined visual and auditory cues. CASES \cite{10.1145/3610910} runs on AI glasses to assist reading by analyzing visual attention and text. Also on AI glasses, OS-1 \cite{10.1145/3659600} leverages an LLM for conversational tutoring, while EMOShip \cite{Zhao_2022} performs emotion analysis. Vinci \cite{huang2025vinci} integrates visual and language understanding to support daily activities. EgoAgent \cite{chen2025egoagentjointpredictiveagent} predicts future states and human poses from egocentric views, and Embodied VideoAgent \cite{fan2025embodiedvideoagentpersistentmemory} enables memory and interaction in dynamic 3D scenes through multimodal fusion and tool invocation. However, these systems are largely tailored to specific tasks and user groups and lack generalization. Existing AI assistants are evolving from single-modality processing to multimodal understanding, integrating diverse functionalities to enhance cross-task generalization and better support everyday life. In real-world settings, AI-assisted tasks are often procedural: they follow ordered steps, target specific goals, and pose substantial comprehension challenges. Simple QA tools or single-modality LLM reasoning are often insufficient. These limitations highlight the research significance of EgoProceAssist.

\textit{\textbf{LLMs and VLMs.}} Recent advances in large language models (LLMs) enable multimodal processing, fueling vision language models (VLMs) that excel at video understanding. These systems usually involve feature alignment and instruction tuning. Improving LLM and VLM capabilities can boost areas such as procedural learning, error detection, and video question answering. Yet, current models mostly use third-person data and underperform on egocentric videos. With the proposal of large-scale egocentric datasets (\textit{e.g.}, Ego4D\cite{grauman2022ego4d}\cite{damen2018scaling}), new LLM and VLM frameworks \cite{zhang2024hcqa}\cite{zhang2025hcqa}\cite{storks2024transparent} have emerged to address this gap.

\textit{\textbf{Egocentric Vision.}} Egocentric video, enabled by head-mounted cameras on task performers, provides essential visual data for AI assistants offering real-time human support. Recent advancements in this field are largely attributable to the release of large-scale annotated datasets \cite{grauman2022ego4d}\cite{damen2020epic}\cite{damen2018scaling}, which facilitate various tasks, including action recognition \cite{zhang2019comprehensive}\cite{kong2022human} and action anticipation \cite{trong2017comprehensive}\cite{rasouli2020deep}. Rapid progress in egocentric vision has laid a critical technological foundation for the development of AI assistants. 

\textit{\textbf{Embodied QA.}} It represents a complex research challenge within embodied intelligence \cite{duan2022survey}\cite{das2018embodied}\cite{fan2024embodied}, requiring active exploration, navigation, and task planning in interactive 3D environments. Unlike egocentric video question answering, which is a fundamental function for egocentric AI assistants and emphasizes passive comprehension and temporal reasoning based on video content, Embodied QA focuses on spatial interaction and real-time decision-making. Consequently, Embodied QA necessitates training within simulated 3D scenes, whereas egocentric video QA relies on annotated video datasets for developing content-specific question-answering capabilities.

\textit{\textbf{Video understanding.}} Long video understanding \cite{miech2019howto100m}\cite{xiao2025videoqa} and egocentric video understanding \cite{nguyen2016recognition} are two important areas in video analysis. Long video understanding focuses on processing extended video content, which can last from several minutes to hours. In contrast, egocentric video understanding deals with videos recorded from a first-person perspective and has unique characteristics and specialized applications compared to general long video analysis. This viewpoint facilitates the adoption of methods such as hand pose estimation \cite{prakash2025synthesizing}, gaze prediction \cite{mazzamuto2025gazing}, and hand-object interaction \cite{bansal2024hoi} to improve recognition performance. It is crucial for building an efficient AI assistant to skillfully leverage the characteristics of the egocentric perspective for video understanding.

\subsection{Three Core Functions and Two Enabling Dimensions}

In the following sections, we focus on several core functions that EgoProceAssist is designed to fulfill, grounded in potential user concerns and anticipated needs in procedural tasks. Consider, for example, furniture assembly, where textual instructions alone may be insufficient and difficult to follow. EgoProceAssist, capable of providing stepwise guidance, detecting omitted or out‑of‑order actions, and promptly responding to task‑related queries, would be of substantial practical value. This requires the assistant to first conduct procedural error detection, enabling timely identification of errors such as incorrect steps and execution mistakes (e.g., use of inappropriate tools). It must also recognize key action sequences from observed video, thereby endowing the system with autonomous procedural learning. By retrieving demonstration videos of similar tasks, the assistant can learn how to complete the task and provide more effective guidance. Timely feedback on the user’s current step further facilitates goal achievement. Finally, the assistant must support continuous procedural question answering and provide timely guidance. On this basis, we regard Egocentric Procedural Error Detection, Egocentric Procedural Learning, and Egocentric Procedural Question Answering as the three core and indispensable functions of EgoProceAssist.

These three functions are intrinsically interdependent rather than purely parallel. Learning correct action sequences from similar videos and comparing them with observed behavior offers an efficient and accurate mechanism for error detection \cite{huang2025modeling}, while simultaneously supplying rich priors for procedural question answering. Conversely, the strong reasoning capabilities of VQA models can improve the reliability of both procedural learning and error detection \cite{flaborea2024prego,plini2025ti}. Moreover, information about incorrect steps obtained via error detection can be used to validate and refine procedural learning and question answering. In subsequent work, we categorize these three fundamental functions according to their functional roles, thereby enabling clearer summarization, classification, and generalization.

While the three core functionalities outlined above define what EgoProceAssist can do, two critical enabling dimensions determine whether the model can be deployed in real-world settings. First, EgoProceAssist must process real-time, streaming video. Only an assistant with low latency and rapid response can provide effective and timely support, but detecting a procedural error five seconds after it occurs offers little practical value. Second, EgoProceAssist must provide proactive guidance and assistance. Existing procedural task-support methods, such as procedural question answering, require users to actively initiate queries. In real-world scenarios where an AI assistant is needed, users often lack the expertise to formulate appropriate questions. Moreover, in urgent error cases, users may be unaware of the problem and thus cannot trigger passive detection or assistance. Consequently, these requirements introduce Real-time and Streaming Video Understanding, and Proactive Interaction in Procedural Contexts as two key enabling dimensions for EgoProceAssist.

\section{Egocentric Procedural Error Detection}
\subsection{Definition}

Unlike general task analysis, procedural tasks such as cooking or assembling require a strict sequence, with each step contingent on the successful completion of previous ones. Mistakes, such as missing, incorrect, or extra steps, can lead to significant errors or hazards. The concept of error detection in procedural tasks was initiated by \cite{sener2022assembly101}, whereas prior work focused on anomaly detection \cite{sultani2018real} or unintentional actions detection \cite{epstein2020oops}, which identify abnormal behaviors based on semantic deviation visible in visual data (\textit{e.g.}, accidents, mistakes). However, for procedural task error detection, correct or incorrect actions are context-dependent; a step that seems plausible in isolation can be erroneous when executed at an improper time order.

Video anomaly detection, which typically operates from a static viewpoint and flags deviations from regular patterns using semantic rules (e.g., identifying falls), differs fundamentally from procedural error detection. Procedural errors are goal-oriented and require long-term context-sensitive reasoning, rendering standard anomaly detection methods insufficient for identifying them.

\begin{figure}[t]
    \centering
    \includegraphics[width=1\linewidth]{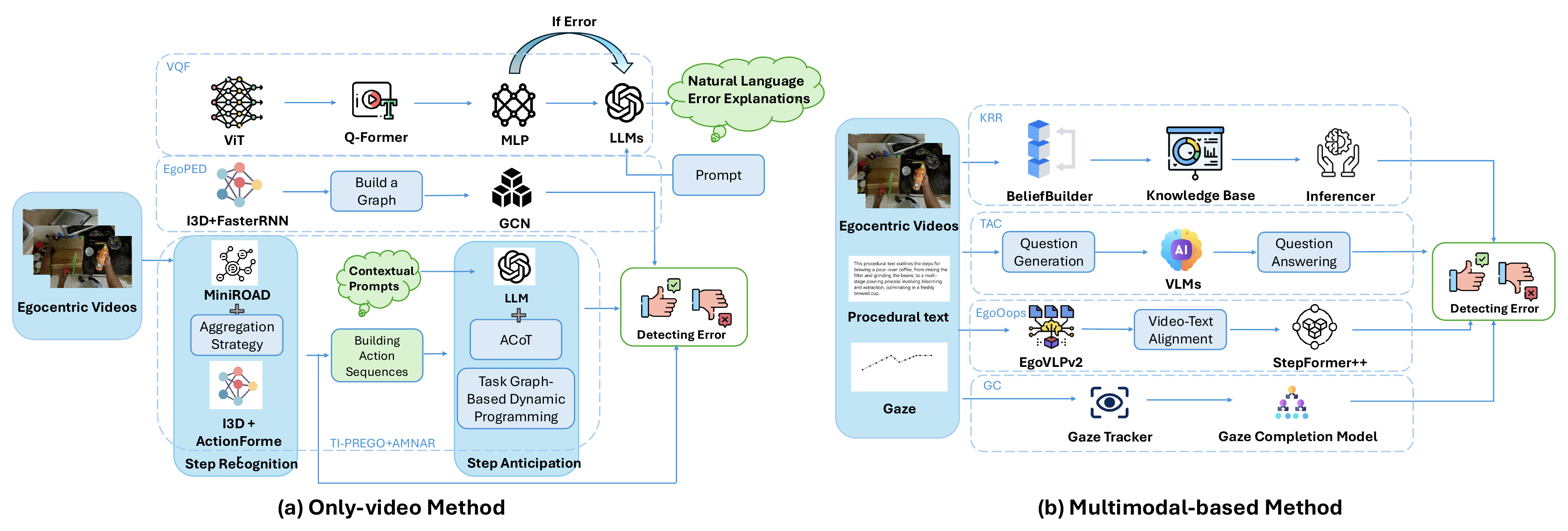}
    \caption{(a) shows the overall framework flowchart summarizing only-video egocentric procedural error detection methods: VQF \cite{patsch2025technical}, EgoPED \cite{lee2024error}, TI-PREGO \cite{plini2025ti}, AMNAR \cite{huang2025modeling}. (b) shows a flowchart summarizing multimodal-based methods: KRR \cite{ding2023every}, TAC \cite{storks2024transparent}, EgoOops \cite{haneji2025egooops}, GC \cite{mazzamuto2025gazing}, where some methods use gaze or procedure-related text as additional input.}
    \label{fig4}%
\end{figure}

As shown in Fig.~\ref{fig4}, current error detection methods differ in their data usage. Only-video methods rely solely on video, while multimodal approaches utilize multiple data types, such as gaze, to achieve higher accuracy, reduce data requirements, and increase practicality. Both techniques help inform the development of real-time egocentric procedural AI assistants. The performance comparison of the various methods is given in Table~\ref{tab1}.

\subsection{Only-video Methods}

Although relying solely on video input, some methods such as Flaborea et al. (PREGO) \cite{flaborea2024prego} and Plini et al. (TI-PREGO) \cite{plini2025ti} leverage the reasoning capabilities of LLMs for contextual next-action prediction, enabling timely responses and the detection of previously unseen errors. The incorporation of Automatic Chain of Thought (ACoT) \cite{zhang2022automatic} reasoning further enhances reasoning ability and improves detection accuracy. However, these approaches are highly dependent on the precision of action recognition and prediction, and may exhibit limited generalization due to the inherent constraints of LLMs. To reduce reliance on prediction accuracy, other methods, such as Huang et al. (AMNAR) \cite{huang2025modeling} and Lee et al. (EgoPED) \cite{lee2024error}, generate multiple dynamic representations of correct actions and detect errors by comparing them. Although they cannot provide rapid prediction-based responses or effectively detect unseen errors, they enable robust error detection across diverse task executions and environmental contexts. These methods can accommodate changes in action distributions, handle scenarios with multiple valid actions, and detect various error types, such as omissions and modifications. To improve error interpretability, Patsch et al. (VQF) \cite{patsch2025technical} propose a Video Q-Former–based end-to-end framework that simultaneously detects sequential and execution errors while providing textual explanations for the identified mistakes.

\begin{table}[h]
 \caption{Performance comparison of representative methods for egocentric procedural error detection, where metric values are averaged across different tasks within the same dataset, the bold text represents the best performance on this metric.}\label{tab1}%
    \centering
            
    \begin{tabular}{cccccccc|c|}
    \toprule
       
          \multicolumn{1}{c|}{ \cellcolor[HTML]{EFEFEF}Method}&   \multicolumn{1}{c|}{ \cellcolor[HTML]{EFEFEF}Years}&   \multicolumn{1}{c|}{\cellcolor[HTML]{EFEFEF}Precision}&   \multicolumn{1}{c|}{ \cellcolor[HTML]{EFEFEF}EDA}&   \multicolumn{1}{c|}{ \cellcolor[HTML]{EFEFEF}AUC}&   \multicolumn{1}{c|}{ \cellcolor[HTML]{EFEFEF}F1}&   \multicolumn{1}{c|}{ \cellcolor[HTML]{EFEFEF}Recall}&   \multicolumn{1}{c|}{ \cellcolor[HTML]{EFEFEF}Acc}&  \multicolumn{1}{c|}{ \cellcolor[HTML]{EFEFEF}Dataset}\\
         \midrule
         KRR \cite{ding2023every}&  arXiv 2023&  63.7&  -&  -&  71.8&  70.6&  86.0& Assemmbly101\\
         \midrule
         \cellcolor{gray!10}EgoPED (MSTCN++) \cite{lee2024error}&  CVPR 2024&  -&  56.7&  54.7&  -&  -&  -& \multirow{7}{*}{HoloAssist}\\
 EgoPED (DiffACT) \cite{lee2024error}& CVPR 2024& -& 69.8& 47.0& -& -& -&\\
 \cellcolor{gray!10}EgoPED (AF) \cite{lee2024error}& CVPR 2024& -& 69.5& 49.1& -& -& -&\\
         VQF \cite{patsch2025technical}&   arXiv 2025&  11.0&  -&  -&  \textbf{55.0}&  21.0&  -& \\
         \cellcolor{gray!10}GC (One-class training) \cite{mazzamuto2025gazing}&  CVPR 2025&  \textbf{14.0}&  -&  \textbf{61.0}&  22.0&  \textbf{59.0}&  -& \\
         GC (Unsupervised training) \cite{mazzamuto2025gazing}&  CVPR 2025&  12.0&  -&  59.0&  18.0&  40.0&  -& \\
         \cellcolor{gray!10}AMNAR \cite{huang2025modeling}&  CVPR 2025&  -&  \textbf{69.9}&  56.5&  -&  -&  -& \\
         \midrule
 PREGO (Oracle+GPT-3.5) \cite{flaborea2024prego}& CVPR 2024& 29.2& -& -& 42.1& 75.8& -&\multirow{7}{*}{Assembly101-O}\\
 \cellcolor{gray!10}PREGO (Oracle+LLAMA) \cite{flaborea2024prego}& CVPR 2024& \textbf{30.7}& -& -& \textbf{46.3}& 94.0& -&\\
 PREGO (OadTR+LLAMA) \cite{flaborea2024prego}& CVPR 2024& 22.1& -& -& 35.8& 94.2& -&\\
 \cellcolor{gray!10}PREGO (MiniRoad+GPT-3.5) \cite{flaborea2024prego}& CVPR 2024& 16.2& -& -& 27.3& 87.5& -&\\
 PREGO (MiniRoad+LLAMA) \cite{flaborea2024prego}& CVPR 2024& 27.8& -& -& 41.8& 84.1& -&\\
\cellcolor{gray!10}TI-PREGO (Oracle+Llama 3.1) \cite{plini2025ti}& CVIU 2025& -& -& -& -& \textbf{97.8}& -&\\
 TI-PREGO (MiniRoad+Llama 3.1) \cite{plini2025ti}& CVIU 2025& -& -& -& -& 97.2& -&\\
 \midrule
 \cellcolor{gray!10}PREGO (Oracle+GPT-3.5) \cite{flaborea2024prego}& CVPR 2024& 9.9& -& -& 17.4& 73.3& -&\multirow{7}{*}{Epic-tent-O}\\
 PREGO (Oracle+LLAMA) \cite{flaborea2024prego}& CVPR 2024& \textbf{10.7}& -& -& \textbf{19.1}& 86.7& -&\\
 \cellcolor{gray!10}PREGO (OadTR+LLAMA) \cite{flaborea2024prego}& CVPR 2024& 9.5& -& -& 17.2& 93.3& -&\\
 PREGO (MiniRoad+GPT-3.5) \cite{flaborea2024prego}& CVPR 2024& 4.3& -& -& 8.0& 66.6& -&\\
 \cellcolor{gray!10}PREGO (MiniRoad+LLAMA) \cite{flaborea2024prego}& CVPR 2024& 8.6& -& -& 12.0& 20.0& -&\\
 TI-PREGO (Oracle+Llama 3.1) \cite{plini2025ti}& CVIU 2025& -& -& -& -& \textbf{100.0}& -&\\
 \cellcolor{gray!10}TI-PREGO (MiniRoad+Llama 3.1) \cite{plini2025ti}& CVIU 2025& -& -& -& -& 93.3& -&\\
 \midrule
 EgoPED (MSTCN++) \cite{lee2024error}& CVPR 2024& -& 48.4& 58.5& -& -& \textbf{74.6}&\multirow{4}{*}{EgoPER}\\
 \cellcolor{gray!10}EgoPED (DiffACT) \cite{lee2024error}& CVPR 2024& -& 49.2& 61.9& -& -& 69.5&\\
 EgoPED (AF) \cite{lee2024error}& CVPR 2024& -& 57.0& 62.0& -& -& 68.5&\\
 \cellcolor{gray!10}AMNAR \cite{huang2025modeling}& CVPR 2025& -& \textbf{64.4}& \textbf{68.5}& -& -& -&\\
 \midrule
 GC (One-class training) \cite{mazzamuto2025gazing}& CVPR 2025& \textbf{37.0}& -& 69.0& \textbf{52.0}& 85.0& -&\multirow{2}{*}{Epic-tent}\\
 \cellcolor{gray!10}GC (Unsupervised training) \cite{mazzamuto2025gazing}& CVPR 2025& 36.0& -& 69.0& 51.0& 85.0& -&\\
  \midrule
 GC (One-class training) \cite{mazzamuto2025gazing}& CVPR 2025& \textbf{18.0}& -& \textbf{63.0}& \textbf{24.0}& \textbf{35.0}& -&\multirow{2}{*}{IndustReal}\\
 \cellcolor{gray!10}GC (Unsupervised training) \cite{mazzamuto2025gazing}& CVPR 2025& 16.0& -& 62.0& 21.0& 33.0& -&\\
 \midrule
 EgoPED \cite{lee2024error}& CVPR 2024& 56.5& 69.8& 54.9& -& -& -&\multirow{2}{*}{CaptainCook4D}\\
 \cellcolor{gray!10}AMNAR \cite{huang2025modeling}& CVPR 2025& \textbf{66.8}& \textbf{72.3}& \textbf{60.2}& -& -& -&\\
 
 \bottomrule
    \end{tabular}
   
\end{table}

\subsection{Multimodal-based Methods}

Besides only-video approaches, numerous methods integrate multimodal data. Wearable devices nowadays are well-suited for collecting various modal information beyond egocentric visual perception, such as gaze, facial micro-expressions, and audio. Gaze reveals user intentions via attentional patterns, while audio conveys social states and environmental cues that are inaccessible to the egocentric view, and vocal characteristics enable inference of users’ emotions and intentions. However, existing error detection methods only utilize information such as text and gaze. How to better utilize data from other modalities warrants further consideration.

Ding et al. (KRR) \cite{ding2023every} construct a dynamically updated knowledge base that formalizes assembly logic into inferable rules, enabling adaptation to complex scenarios by comparing current and historical actions. Other approaches like Storks et al. (TAC) \cite{storks2024transparent} and Haneji et al. (EgoOops) \cite{haneji2025egooops} integrate textual data to provide transparent and interpretable error detection. However, these methods generally rely on fine-grained dataset annotations, and their performance degrades when generalizing to unseen domains. Meanwhile, video-text-based approaches often suffer from the loss of fine-grained details. Mazzamuto et al. (GC) \cite{mazzamuto2025gazing} addresses these limitations by employing analysis of users’ eye movements. Based on gaze prediction, it does not require access to the full video and can respond rapidly, even to previously unseen error types.

\begin{table}
\caption{Summarize the representative datasets in egocentric procedural error detection, Ego means egocentric view and Exo means third-person view.}
    \centering

     \begin{tabular}{cccccccc}
     \toprule
        \multicolumn{1}{c|}{\cellcolor[HTML]{EFEFEF}Dataset}&   \multicolumn{1}{c|}{ \cellcolor[HTML]{EFEFEF}Years}&   \multicolumn{1}{c|}{ \cellcolor[HTML]{EFEFEF}Duration}&   \multicolumn{1}{c|}{\cellcolor[HTML]{EFEFEF}Videos}&   \multicolumn{1}{c|}{ \cellcolor[HTML]{EFEFEF}Segments}&   \multicolumn{1}{c|}{ \cellcolor[HTML]{EFEFEF}Activity}&   \multicolumn{1}{c|}{ \cellcolor[HTML]{EFEFEF}Domain}&  \multicolumn{1}{c}{ \cellcolor[HTML]{EFEFEF}View} \\
         \midrule
          EPIC-KITCHENS \cite{damen2020epic}&  ECCV 2018&  200h&  700&  90k&  -&Cooking& Ego \\
          Epic-tent \cite{jang2019epic}& ICCV 2019& 5.4h& 72& 921& 38&Outdoor&Ego \\
          EgoCom \cite{northcutt2020egocom}&  TPAMI 2020& 38.5h& 70& -& -&Communication&Ego \\
         Assembly101 \cite{sener2022assembly101}&  CVPR 2022&  -&  4321&  1M&  15&Assembly& Ego \\
         Ego4D \cite{grauman2022ego4d}&  CVPR 2022&  3670h&  -&  -&  100+&Mixed& Ego \\
         HoloAssist \cite{wang2023holoassist}&  ICCV 2023&  166h&  -&  -&  20&Mixed& Ego \\
         Ego-Exo4D \cite{grauman2024ego}&  CVPR 2024&  1286h&  -&  -&  123&Mixed& Ego+Exo \\
         IKEA Manuals at Work \cite{liu2024ikea}& NeurIPS 2024&  10h&  98&  1120&  -&Assembly& Ego+Exo \\
         EgoOops \cite{haneji2025egooops}& ICCV 2025& 6.8h& 50& 538& 5&Mixed&Ego \\
         EgoPER \cite{lee2024error}& CVPR 2024& 28h& 386& -& 5&Cooking&Ego \\
         IndustReal \cite{schoonbeek2024industreal}& WACV 2024& 5.8h& 84& 9273& 75&Industrial&Ego \\ 
         CaptainCook4D \cite{peddi2024captaincook4d}& NeurIPS 2024& 94.5h& 384& 5.8k& -&Cooking&Ego \\   
   \bottomrule
      \end{tabular}

\label{tab2}%

\end{table}

\subsection{Datasets}

This section reviews egocentric datasets relevant to egocentric procedural error detection, as shown in Table~\ref{tab2}, with a focus on datasets that involve procedural activities. We summarize and analyze these datasets by scale and data type.

\textbf{\textit{EPIC-Kitchens}} \cite{damen2020epic} is a large-scale egocentric dataset of 32 participants cooking in diverse home kitchens, totaling 55 hours of unscripted video with audio. The dataset includes rich annotations spanning action categories, temporal boundaries, and participant metadata, but excludes segments with errors.
\textbf{\textit{EPIC-Tent}} \cite{jang2019epic} features outdoor tent-pitching videos from 24 users wearing head-mounted cameras, addressing non-rigid object manipulation scenarios. It uniquely includes participant-provided subjective uncertainty scores, and contains both procedural and execution errors.
\textbf{\textit{EgoCom}} \cite{northcutt2020egocom} comprises 38.5 hours of synchronized binaural audio and egocentric video, with 20,000 timestamped transcriptions and speaker labels from 34 speakers, providing interactive visual and conversational data, including implicit cues absent in third-person datasets, but excludes segments with errors.
\textbf{\textit{Assembly 101}} \cite{sener2022assembly101} contains 4,321 videos of users assembling and disassembling 101 types of toy vehicles, allowing diverse action sequences, step errors, and corrections. It includes annotations for over 100k coarse- and 1M fine-grained action segments, as well as 18M 3D hand poses. Unlike other datasets, it primarily consists of procedural errors.
\textbf{\textit{Ego4D}} \cite{grauman2022ego4d}  comprises over 3,670 hours of daily activity videos from 74 locations in 9 countries, captured with seven types of head-mounted cameras. It includes diverse environments and provides multimodal data, including 3D scans, audio, gaze data, and stereo recordings, but excludes segments with errors.
\textbf{\textit{HoloAssist}} \cite{wang2023holoassist} is tailored for AI assistant training in physical tasks, featuring 166 hours of data from 350 performer-instructor pairs across 20 object manipulation categories. It provides 414 coarse- and 1,887 fine-grained action classes, with manual annotations for actions, error types, and dialogues. Timely error corrections and proactive interventions are included, along with both procedural errors and execution errors.
\textbf{\textit{Ego-Exo4D}} \cite{grauman2024ego} features over 1,400 hours of paired egocentric and third-person videos from nearly 800 skilled participants across diverse scenarios, including sports, music, dance, and bicycle repair, capturing both egocentric and contextual views. Simultaneously encompasses both procedural and execution errors.
\textbf{\textit{IKEA Manuals at Work}} \cite{liu2024ikea} targets furniture assembly with complex scenes, aligning manual instructions and videos for 4D localization of 3D assembly processes. It includes 3D models for 36 items, 98 real assembly videos, and detailed step-level spatiotemporal annotations, but excludes segments with errors.
\textbf{\textit{EgoOops}} \cite{haneji2025egooops} is dedicated to error detection, spanning diverse tasks such as circuit assembly, chemistry experiments, color mixing, block construction, and paper crafts. Unlike prior datasets focused on assembly or cooking, EgoOops offers both correct and incorrect demonstrations with detailed fine-grained annotations, and encompasses both procedural and execution errors.
\textbf{\textit{EgoPER}} \cite{lee2024error} comprises 386 kitchen cooking videos (28 hours), with 213 showing correct and 173 showing erroneous executions. It features a manually constructed recipe task graph covering all correct and incorrect paths for script-based video generation, and encompasses both procedural and execution errors.
\textbf{\textit{IndustReal}} \cite{schoonbeek2024industreal} features 84 videos (5.8 hours) of 27 participants assembling 3D-printed toy cars with HoloLens 2. It covers both assembly and maintenance, documenting 724 correct and 38 error steps across 38 error types, including procedural and execution errors.
\textbf{\textit{CaptainCook4D}} \cite{peddi2024captaincook4d} offers 384 kitchen cooking videos (94.5 hours) across 24 recipes, capturing both procedural and execution errors. It includes 5.3K step-level and 10K fine-grained action annotations.

\subsection{Evaluation}
 
Current egocentric procedural error detection is commonly evaluated using F1 score \cite{ding2023every}\cite{flaborea2024prego}, Accuracy\cite{ding2023every}, per-class Precision \cite{flaborea2024prego} and Recall \cite{flaborea2024prego}. Additional metrics such as Error Detection Accuracy (EDA) \cite{huang2025modeling}, Area Under the Curve (AUC) \cite{huang2025modeling}, and Edit Distance are also employed in relevant studies.

Precision denotes the proportion of true positive error detections among all predicted errors, reflecting the reliability of the model’s error predictions. Recall represents the ratio of true positive error detections to all actual error instances, indicating the model’s completeness in identifying errors. Accuracy is the proportion of correctly classified instances (both erroneous and normal) among all samples, reflecting the overall correctness of the model. The F1 scores or F1@$\mathcal{\tau}$ assess segmentation by calculating the Intersection over Union (IoU) between predicted and ground-truth segments at a threshold $\mathcal{\tau}$/100. Segments with IoU above the threshold are counted as true positives, while additional overlapping predictions are considered false positives. The F1 score is computed as the harmonic mean of precision and recall under these criteria: 
\begin{equation}
    \mathrm{F1}=2\cdot\frac{\mathrm{precision}*\mathrm{recall}}{\mathrm{precision}+\mathrm{recall}}.
\end{equation}
Typically, $\mathcal{\tau}$ is chosen from the set \{10, 25, 50\}. 

Semantic similarity of error explanations is evaluated using BLEU for n-gram overlap, ROUGE-L for longest common subsequence, and CIDEr for TF-IDF weighted similarity. Omission Accuracy (O-Acc) quantifies the proportion of detected ground-truth omission errors. Omission Intersection over Union (O-IoU) measures the overlap between predicted and ground-truth omission errors:
\begin{equation}
    \mathrm{O-IoU}=\frac{|GT_o\cap PD_o|}{|GT_o\cup PD_o|},
\end{equation}
where $GT_o$ is the set of ground-truth omission errors and $PD_o$ is the set of predicted omission errors.

Error Detection Accuracy (EDA) is the proportion of segments correctly classified as erroneous or normal, indicating overall segment-level performance in error detection. Area Under the Curve (AUC) quantifies the model’s discriminative ability by integrating the true positive and false positive rates across varying thresholds and is commonly used in binary classification. Edit Distance measures the dissimilarity between predicted and ground-truth step sequences, enabling assessment of omissions and ordering errors.

\section{Egocentric Procedural Learning}

\subsection{Definition}

Procedure learning entails identifying key steps and their logical order while filtering out irrelevant or redundant actions, given the variability in task execution sequences. For AI assistants, autonomously acquiring step sequences facilitates effective error detection and enables stepwise task guidance.

Action recognition \cite{kong2022human} classifies the main action in a segment without considering temporal boundaries. Action segmentation \cite{ding2023temporal} assigns frame-level action labels to untrimmed videos, identifying both action types and boundaries. In contrast, procedure learning segments and analyzes multiple videos to identify and sequence key steps for completing a specific task.

Recent advances in egocentric procedural learning rely on varying levels of annotation, with mainstream methods categorized as self-supervised, unsupervised, or weakly-supervised. Organizing these approaches by supervision level provides a clear analytical framework. The following sections review representative works in each category. The performance comparison of the various methods is given in Table~\ref{tab3}.

\begin{table}
 \caption{Performance comparison of representative methods of egocentric procedural learning, where metric values are averaged across different tasks within the same dataset, the bold text represents the best performance on this metric.}
    \centering
    \begin{tabular}{cccccccc|c|}
    \toprule
         \multicolumn{1}{c|}{ \cellcolor[HTML]{EFEFEF}Method}&  \multicolumn{1}{c|}{ \cellcolor[HTML]{EFEFEF}Years}&  \multicolumn{1}{c|}{\cellcolor[HTML]{EFEFEF}Precision}&  \multicolumn{1}{c|}{\cellcolor[HTML]{EFEFEF}IoU}&  \multicolumn{1}{c|}{ \cellcolor[HTML]{EFEFEF}F1}&  \multicolumn{1}{c|}{ \cellcolor[HTML]{EFEFEF}Recall} &  \multicolumn{1}{c|}{ \cellcolor[HTML]{EFEFEF}mIoU}& \multicolumn{1}{c|}{ \cellcolor[HTML]{EFEFEF}Jaccard}&  \multicolumn{1}{c|}{ \cellcolor[HTML]{EFEFEF}Dataset}\\
         \midrule
         ULNI \cite{alayrac2016unsupervised}&  CVPR 2016&  76.0&  -&  \textbf{46.0}&  67.0 & -&-&  \multirow{2}{*}{YouTube}\\
         \cellcolor{gray!10}ULAC \cite{kukleva2019unsupervised}&  CVPR 2019&  -&  -&  28.3&  - & -&-&  \\
         
          \midrule
 ProcNets-NMS \cite{zhou2018towards}& AAAI 2018& 30.4& -& 33.4&  37.1& 33.9&47.6&\multirow{2}{*}{YouCook2}\\
 \cellcolor{gray!10}ProcNets-LSTM \cite{zhou2018towards}& AAAI 2018& -& -& -& -& \textbf{37.0}& \textbf{50.6}&\\
  \midrule
         
         \cellcolor{gray!10}ULAC \cite{kukleva2019unsupervised}&  CVPR 2019&  -&  -&  26.4&  - & -&-&  \multirow{3}{*}{Breakfast}\\
 SPS-FL \cite{naing2020procedure}& BMVC 2020& -& -& 45.4& - & -&-& \\
 \cellcolor{gray!10}SPS-LC \cite{naing2020procedure}& BMVC 2020& -& -& \textbf{53.2}& - & -&-& \\
 \midrule
 SPS-FL \cite{naing2020procedure}& BMVC 2020& 39.6& -& 48.1& 64.2 & -&-& \multirow{2}{*}{Inria}\\
 \cellcolor{gray!10}SPS-LC \cite{naing2020procedure}& BMVC 2020& \textbf{47.4}& -& \textbf{56.0}& \textbf{72.8} & -&-& \\

         \midrule
         JointSeqFL \cite{elhamifar2019unsupervised} &  ICCV 2019&  -&  -&  28.6&  - & -&-&  \multirow{6}{*}{ProceL}\\
          \cellcolor{gray!10}CnC \cite{bansal2022my}& ECCV 2022& 20.7& -& 21.6& 22.6& -& -&\\
 STEPs \cite{shah2023steps}& ICCV 2023& 23.5& -& 24.9& 26.7& -& -&\\
         \cellcolor{gray!10}GPL \cite{bansal2024united}&  WACV 2024&  22.4&  &  23.4&  24.5 &     -& -&  \\
 OPEL \cite{chowdhury2024opel}& NeurIPS 2024& 33.6& 17.9& 34.9& 36.3 & -&-& \\
 \cellcolor{gray!10}RGWOT \cite{mahmood2025procedure}& arXiv 2025& \textbf{42.2}& \textbf{29.4}& \textbf{44.3}& \textbf{46.7} & -&-& \\

 \midrule
  CnC \cite{bansal2022my}& ECCV 2022& 22.8& -& 22.6& 22.5& -& -&\multirow{5}{*}{CrossTask}\\
 \cellcolor{gray!10}STEPs \cite{shah2023steps}& ICCV 2023& 26.2& -& 25.9& 25.8& -& -&\\
 GPL \cite{bansal2024united}& WACV 2024& 24.9& -& 24.5&  24.1& -&-& \\
\cellcolor{gray!10} OPEL \cite{chowdhury2024opel}& NeurIPS 2024& 35.6& 16.9& 35.1& 34.8 & -&-& \\
 RGWOT \cite{mahmood2025procedure}& arXiv 2025& \textbf{40.4}& \textbf{26.3}& \textbf{40.4}& \textbf{40.7} & -&-& \\

 \midrule
   \cellcolor{gray!10}CnC \cite{bansal2022my}& ECCV 2022& -& 10.7& 22.0& -& -& -&\multirow{4}{*}{EgoProceL}\\
 GPL \cite{bansal2024united}& WACV 2024& -& 13.9& 25.6& -& -& -&\\
 \cellcolor{gray!10}OPEL \cite{chowdhury2024opel}& NeurIPS 2024& -& 16.3& 32.0& - & -&-& \\
 RGWOT \cite{mahmood2025procedure}& arXiv 2025& -& \textbf{31.2}& \textbf{46.8}& - & -&-& \\

  \bottomrule

    \end{tabular}
   
   \label{tab3}%
\end{table}

\begin{figure}
    \centering
    \includegraphics[width=1\linewidth]{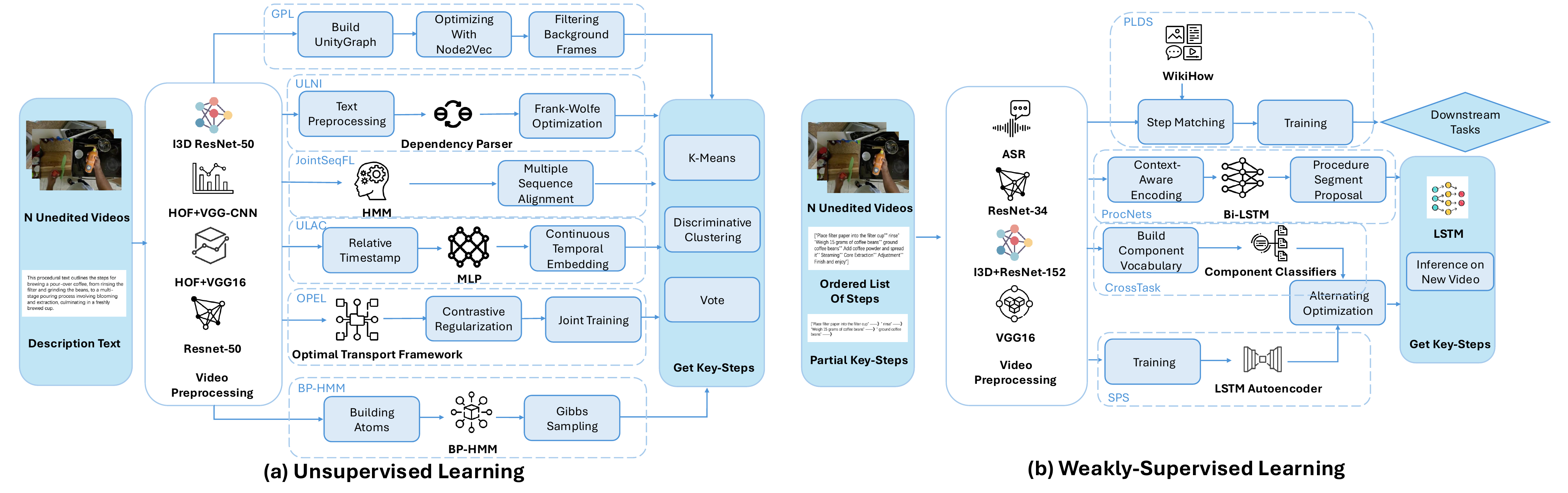}
    \caption{(a) shows a flowchart summarizing unsupervised learning methods: GPL \cite{bansal2024united}, ULNI \cite{alayrac2016unsupervised}, JointSeqFL \cite{elhamifar2019unsupervised}, ULAC \cite{kukleva2019unsupervised}, OPEL \cite{chowdhury2024opel}, BP-HMM \cite{sener2015unsupervised}, (b) shows a flowchart summarizing weakly-supervised learning methods: PLDS \cite{lin2022learning}, ProNets \cite{zhou2018towards}, CrossTask \cite{zhukov2019cross}, SPS \cite{naing2020procedure}.}
    \label{fig5}%
\end{figure}

\subsection{ Unsupervised Learning}

Unsupervised learning operates without manual labels, leveraging raw data, ideal for annotation-scarce settings. Unlike supervised methods, it avoids label-induced biases and exploits latent information in unannotated data. In procedural learning, unsupervised approaches are mainly utilized for clustering. A flowchart is shown in Fig.~\ref{fig5}.

Clustering partitions data into groups of high intra-cluster similarity and low inter-cluster similarity, utilizing the data’s inherent distribution without relying on predefined labels. Given a dataset $X=\{x_1,x_2,...,x_n\}$, where $x_i\in\mathbb{R}^d$, the goal is to find a mapping:
\begin{equation}
    C:X\to\{1,2,...,k\},
\end{equation}
where $k$ is the number of clusters, and $C(x_i)$ denotes the cluster label of sample $x_i$. Clustering is typically achieved by optimizing certain objective functions; for example, K-means minimizes the Sum of Squared Errors (SSE) within clusters:
\begin{equation}
    \mathrm{SSE}=\sum_{i=1}^k\sum_{x\in C_i}\|x-\mu_i\|^2,
\end{equation}
where $\mu_i$ is the center of cluster $C_i$.

Bansal et al. (GPL) \cite{bansal2024united} introduce a graph-based framework with K-means clustering for unsupervised key-step discovery using spatiotemporal and semantic relations. And Elhamifar et al. (JointSeqFL) \cite{elhamifar2019unsupervised} employ HMMs and optimization to select representative sub-activity sequences. Both can effectively filter background frames, handle repeated key-steps, as well as missing or additional actions. However, they rely on the semantic consistency of similar objects and actions, resulting in limited performance for detecting steps involving highly diverse objects or lacking hand-object interactions. In addition, its time complexity grows exponentially with the number of videos. Sener et al. (BP-HMM) \cite{sener2015unsupervised} automatically discover and parse semantic steps in large video collections, generating textual descriptions, and also face the same limitations. Alayrac et al. (ULNI) \cite{alayrac2016unsupervised} performed text clustering to guide discriminative video clustering for key-step identification. It can exploit the complementarity of the two modalities to resolve ambiguities arising from a single modality, and demonstrates strong performance on tasks with highly discriminative textual descriptions, such as "changing a tire." To further enable unsupervised learning for activities with unknown categories, Kukleva et al. (ULAC) \cite{kukleva2019unsupervised} learn frame embeddings with an MLP and order key-steps based on the Viterbi algorithm, but it relies excessively on the relative temporal positions of frames, resulting in poor adaptability to action sequences with non-monotonic temporal structures. Chowdhury et al. (OPEL) \cite{chowdhury2024opel} use Optimal Transport theory for video alignment and key-step clustering, thereby avoiding the need for manual annotation, however, it assumes that the objects involved are similar when executing the same key step; large object discrepancies can therefore lead to alignment errors.

\subsection{ Weakly-Supervised Learning}

Weakly-supervised methods use limited annotations (\textit{e.g.}, video-level tags), thereby reducing labeling costs compared to full supervision. They offer greater accuracy and a narrower search space than unsupervised methods, and avoid proxy task limitations and pseudo-label errors common in self-supervised approaches.

Lin et al. (PLDS) \cite{lin2022learning}\cite{anthonio2020wikihowtoimprove} apply distant supervision, generating pseudo-labels by aligning ASR transcripts with external text. Still, it relies heavily on a high-quality textual knowledge base; consequently, transcription noise in speech inputs and insufficient coverage of the knowledge base can adversely affect its performance. To reduce annotation cost, Zhou et al. (ProcNets) \cite{zhou2018towards} train using only temporal boundaries, Zhukov et al. (CrossTask) \cite{zhukov2019cross} use shared component classifiers and ordered step lists with narrations for cross-task supervision without temporal labels, and it can further parse previously unseen tasks. Also, Naing et al. (SPS) \cite{naing2020procedure} infer and localize unannotated key steps from partial summaries, framing step completion as joint representation learning and greedy search. Peirone et al. (HiERO) \cite{peirone2025hiero} use a hierarchical graph structure, where a temporal encoder aggregates local temporal information and a function-aware decoder clusters functionally similar segments, and training with video-narration alignment and functional thread losses, the model learns features without task-specific training, achieves strong performance on egocentric video-text alignment and zero-shot procedural learning tasks.

\begin{figure}
    \centering
    \includegraphics[width=0.6\linewidth]{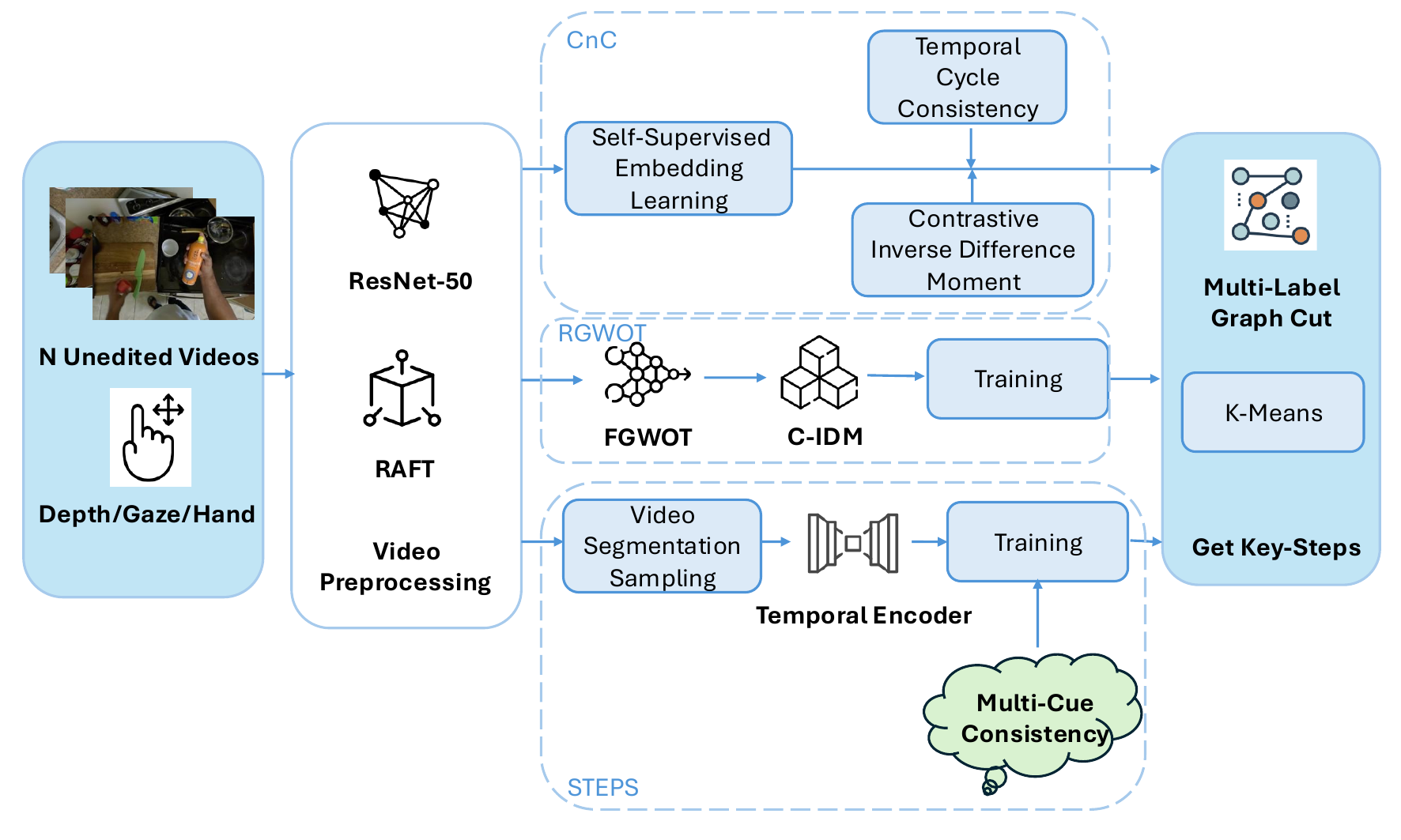}
    \caption{A flowchart summarizing self-supervised egocentric procedural learning methods: CnC\cite{bansal2022my}, RGWOT\cite{mahmood2025procedure} and STEPS\cite{shah2023steps}.}
    \label{fig6}%
\end{figure}

\subsection{Self-Supervised Learning}

Shown in Fig.~\ref{fig6}, Self-supervised learning creates auxiliary tasks to generate supervisory signals from unlabeled data. As a subset of unsupervised learning, it reduces human annotation effort and enhances generalization for large-scale training.

Bansal et al. (CnC) \cite{bansal2022my} use temporal cycle consistency to learn correspondences and generate pseudo-labels, but rely on the temporal correspondence of key-steps across videos, making it vulnerable to repeated and redundant frames, also ineffective for non-monotonic sequences. To address this issue, Mahmood et al. (RGWOT) \cite{mahmood2025procedure} apply FGWOT with a contrastive loss to align frame sequences and incorporate temporal and structural priors, achieving better performance. Further, designing unimodal temporal and cross-modal contrastive losses, Shah et al. (STEPS) \cite{shah2023steps} do not require video alignment or backbone fine-tuning, support multi-sensor modalities on AR devices, and maintain stable performance in few-shot scenarios.

\begin{table}
\caption{Summarize the representative datasets in egocentric procedural learning, Ego means egocentric view, and Exo means third-person view.}
    \centering
    \begin{tabular}{cccccccccc}
    \toprule
         \multicolumn{1}{c|}{\cellcolor[HTML]{EFEFEF}Dataset}&  \multicolumn{1}{c|}{\cellcolor[HTML]{EFEFEF}Years}&  \multicolumn{1}{c|}{\cellcolor[HTML]{EFEFEF}Duration}&  \multicolumn{1}{c|}{\cellcolor[HTML]{EFEFEF}Videos}&  \multicolumn{1}{c|}{\cellcolor[HTML]{EFEFEF}Segments}&  \multicolumn{1}{c|}{\cellcolor[HTML]{EFEFEF}Tasks}&  \multicolumn{1}{c|}{\cellcolor[HTML]{EFEFEF}Steps}&  \multicolumn{1}{c|}{\cellcolor[HTML]{EFEFEF}Domain}& \multicolumn{1}{c|}{\cellcolor[HTML]{EFEFEF}View} &\multicolumn{1}{c}{\cellcolor[HTML]{EFEFEF}Step annotations}\\
         \midrule
         Breakfast \cite{kuehne2014language}&  CVPR 2014&  77h&  -&  -&  10&  48&  Cooking& Exo &\makecell{\ding{51}}\\
         Inria \cite{alayrac2016unsupervised}&  CVPR 2016&  5h&  150&  800k&  5&  -&  Mixed&  Exo&\makecell{\ding{51}}\\
         YouCook2 \cite{zhou2018towards}&  AAAI 2018&  176h&  2000&  14k&  89&  15k&  Cooking& Exo &\makecell{\ding{51}}\\
         COIN \cite{tang2019coin}&  CVPR 2019&  476h&  11827&  46354&  180&  -&  Mixed& Exo &\makecell{\ding{51}}\\
         HowTo100M \cite{miech2019howto100m}&  ICCV 2019&  134472h&  1.22M&  136M&  23611&  -&  Mixed& Exo &\makecell{\ding{55}}\\
         CrossTask \cite{zhukov2019cross}&  CVPR 2019&  376h&  4700&  133&  83&  705&  Mixed& Exo &\makecell{\ding{51}}\\
         ProceL \cite{elhamifar2019unsupervised}&  ICCV 2019&  47.3h&  720&  720&  12&  -&  Mixed& Exo &\makecell{\ding{51}}\\
         EgoProceL \cite{bansal2022my}&  ECCV 2022&  62h&  -&  720&  16&  8.7Avg&  Mixed& Ego &\makecell{\ding{51}}\\
   \bottomrule
    \end{tabular}
    
    \label{tab4}%

\end{table}

\subsection{ Datasets}
 
For procedural error detection, various egocentric datasets have been introduced, covering scenarios such as cooking \cite{damen2020epic}\cite{peddi2024captaincook4d}, assembly \cite{sener2022assembly101}\cite{liu2024ikea}, outdoor tent pitching \cite{jang2019epic}, and specialized domains like chemical experiments \cite{haneji2025egooops}. Several additional datasets are also prevalent in procedural learning research, as shown in Table~\ref{tab4}.

\textbf{\textit{Breakfast}} \cite{kuehne2014language} includes 77 hours of video for 10 cooking activities by 52 participants in 18 kitchens. Each video averages 5.2 unique actions and 6.9 action segments, with 10\% of actions repeated.
\textbf{\textit{Inria}} \cite{alayrac2016unsupervised} is an instructional dataset of 150 videos (30 per task) spanning five tasks, with 800,000 frames and comprehensive annotations.
\textbf{\textit{YouCook2}} \cite{zhou2018towards} contains 2,000 untrimmed videos featuring 89 global cooking recipes, averaging 22 videos per recipe and encompassing diverse cooking styles. 
\textbf{\textit{COIN}} \cite{tang2019coin} contains 11,827 YouTube videos across 180 tasks in 12 daily domains, totaling 476 hours and 46,354 annotated segments, with an average video length of 2.36 minutes.
\textbf{\textit{HowTo100M}} \cite{miech2019howto100m} includes 1.22 million instructional videos, covering 23,000+ visual tasks and 136 million clips, each with narrations and auto-downloaded subtitles from YouTube.
\textbf{\textit{CrossTask}} \cite{zhukov2019cross} consists of instructional videos for 83 tasks (\textit{e.g.}, making a bread, preparing a latte, building a stool), each annotated with an ordered, human-written list of steps.
\textbf{\textit{ProceL}} \cite{elhamifar2019unsupervised} contains 47.3 hours of annotated instructional video from 720 clips across 12 tasks (\~60 per task). Five tasks overlap with Inria \cite{alayrac2016unsupervised}, while seven are new; videos are sourced from YouTube and trimmed for relevance.
\textbf{\textit{EgoProceL}} \cite{bansal2022my} comprises 62 hours of egocentric video across 16 domains, emphasizing key steps for task completion rather than every action. Key step orders may vary across videos.

\subsection{Evaluation}

Evaluation in procedural learning typically employs metrics such as F1 score, Precision, and Recall \cite{bansal2024united}\cite{alayrac2016unsupervised}, especially for error detection. Other approaches may utilize Intersection over Union (IoU) \cite{chowdhury2024opel}, the frame-level metric MoF \cite{kukleva2019unsupervised}, the Hungarian algorithm, MidH, and the Jaccard index \cite{zhou2018towards}.

The Hungarian algorithm can generate a one-to-one mapping between ground truth and predicted values. IoU is calculated using the formula: 
\begin{equation}
    \mathrm{IoU}=\frac{1}{A}\sum_a|GT_a\cap D_a|/|GT_a\cup D_a|,
\end{equation}
where $GT_a$ and $D_a$ represent the ground truth frame set and the predicted frame set for action $a$, respectively.

MoF is the percentage of frames with correctly predicted action labels. It is defined as
\begin{equation}
    { \mathrm { M o F } } = { \frac { \# { \mathrm { o f ~ c o r r e c t ~ f r a m e s } } } { \# { \mathrm { o f ~ a l l ~ f r a m e s } } } }, 
\end{equation}
where \# represents "number".

MidH measures the percentage of predicted action segments whose midpoints fall within a ground-truth action segment. Some methods employ the Jaccard score, which quantifies evaluation by computing the maximum intersection between predicted proposals and ground truth segments for each video, averaging these per video, and then averaging across all videos to obtain the final metric.

These metrics often yield high scores for models that cluster frames into a single group, since the majority key-step typically corresponds to background frames in ground truth. In untrimmed procedural videos, this background predominance can artificially inflate evaluation results. Shen et al. \cite{shen2021learning} examined this issue using the MoF metric, which, as noted by \cite{kukleva2019unsupervised}, is unsuitable for imbalanced datasets. Bansal et al. \cite{bansal2022my} propose computing frame-level scores per key-step and averaging over all key-steps, penalizing uneven performance, such as assigning all frames to a single key-step. This protocol yields lower scores across all methods.

\section{Egocentric Procedural Question Answering}

\subsection{Definiton}

An AI assistant with autonomous procedural learning can identify key steps and provide real-time responses to user queries (\textit{e.g.}, tool location or selection) by reasoning over long-term egocentric videos, thereby enhancing task guidance in daily life.

Video Question Answering (VideoQA) requires models to answer questions about video content using spatial and temporal information to connect computer vision and NLP. Egocentric VideoQA differs from third-person VideoQA, as it contains questions about the camera wearer and features rapid scene changes, diverse actions, and complex long-term dependencies. Consequently, third-person models often underperform in egocentric settings.

\begin{figure}
    \centering
    \includegraphics[width=1\linewidth]{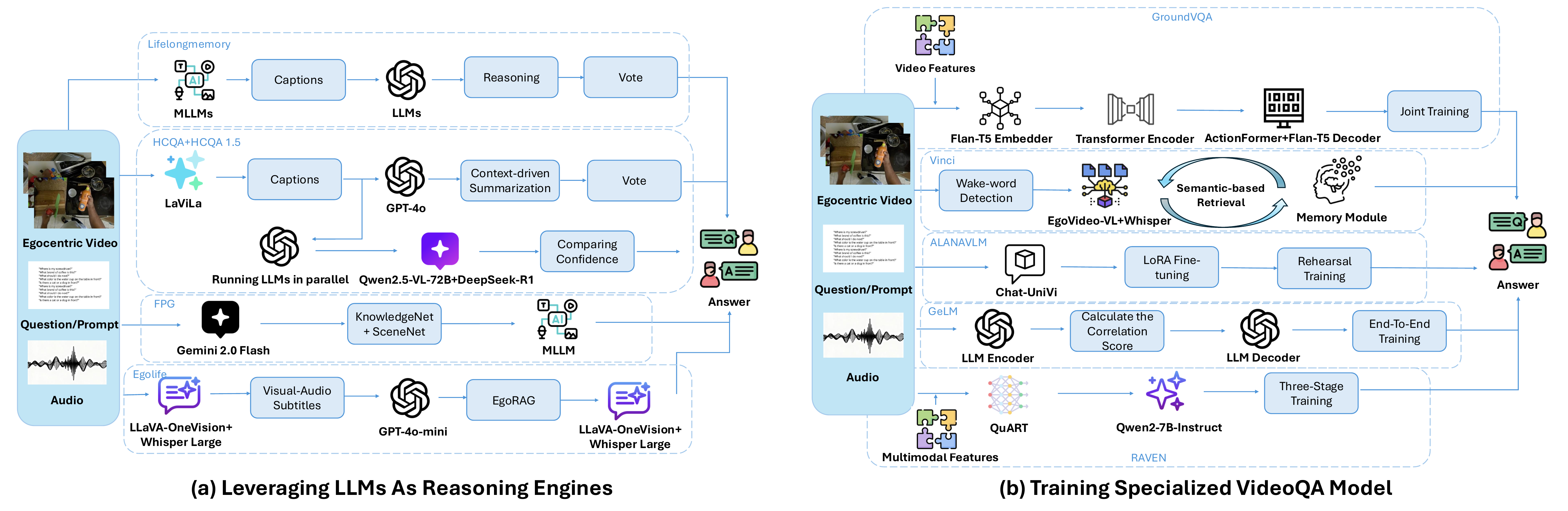}
    \caption{(a) shows the different methods that using LLMs as the main inference engine to obtaining answers: Lifelongmemory \cite{wang2023lifelongmemory}, HCQA \cite{zhang2024hcqa}, HCQA 1.5 \cite{zhang2025hcqa}, FPG \cite{taluzzi2025pixels}, Egolife \cite{yang2025egolife}, and (b) shows the different methods that training specialized videoQA model to obtaining answers: GroundVQA \cite{di2024grounded}, Vinci \cite{huang2025vinci}, ALANAVLM \cite{suglia2024alanavlm}, GeLM \cite{chen2025grounded}, RAVEN \cite{biswas2025raven}.}
    \label{fig7}%
\end{figure}

VideoQA has evolved from “CNN+RNN” to “CNN/ViT+BERT” and now to the “CLIP+LLM” stage \cite{xiao2025videoqa}, leveraging cross-modal encoders (\textit{e.g.}, CLIP-ViT \cite{zeng2025multimodal}\cite{yang2024clip}) and frozen LLMs (\textit{e.g.}, Flan-T5, LLaMA). Current methods either use LLMs as fixed reasoning engines with processed video inputs or train dedicated egocentric VideoQA models. Leading models now approach human-level performance. Subsequent sections review these approaches by category, and the flowchart is shown in Fig.~\ref{fig7}. The performance comparison of the various methods is given in Table~\ref{tab5}.

\begin{table}
\caption{Performance comparison of representative methods of egocentric video question answering, where metric values are averaged across different tasks within the same dataset. The bold text represents the best performance on this metric.}
    \centering
    \begin{tabular}{cccc|c|}
    \toprule
         \multicolumn{1}{c|}{\cellcolor[HTML]{EFEFEF}Method}& \multicolumn{1}{c|}{\cellcolor[HTML]{EFEFEF}Year} &  \multicolumn{1}{c|}{\cellcolor[HTML]{EFEFEF}Accuracy}&  \multicolumn{1}{c|}{\cellcolor[HTML]{EFEFEF}QA Score (0-10)}&  \multicolumn{1}{c|}{\cellcolor[HTML]{EFEFEF}Benchmark}\\
         \midrule
         
         Lifelong (Claude-3-Haiku) \cite{wang2023lifelongmemory}&  arXiv 2023&  55.2&  -&  \multirow{10}{*}{EgoSchema}\\
         \cellcolor{gray!10}Lifelong (GPT-4) \cite{wang2023lifelongmemory}&  arXiv 2023&  62.1&  -&  \\
         Lifelong (GPT-4o) \cite{wang2023lifelongmemory}&  arXiv 2023&  64.6&  -&  \\
 \cellcolor{gray!10}Lifelong (Llama3-8B) \cite{wang2023lifelongmemory}& arXiv 2023& 60.4& -& \\
 Lifelong (GPT-3.5 ) \cite{wang2023lifelongmemory}& arXiv 2023& 64.0& -& \\
 \cellcolor{gray!10}HCQA \cite{zhang2024hcqa}& arXiv 2024& 75.0& -& \\
  MM-Ego \cite{yemmego}& ICLR 2025& 69.0& -&\\
 \cellcolor{gray!10}HCQA 1.5 \cite{zhang2025hcqa}& arXiv 2025& \textbf{77.3}& -& \\

 EgoGPT (EgoIT) \cite{yang2025egolife}&  CVPR 2025&  73.2&  -&  \\
         \cellcolor{gray!10}EgoGPT (EgoIT+D1) \cite{yang2025egolife}&  CVPR 2025&  75.4&  -&  \\
  \midrule
  LLaVA-OneVision \cite{lillava}& TMLR 2025& 47.3& -& \multirow{2}{*}{EgoMemoria}\\
 \cellcolor{gray!10} MM-Ego \cite{yemmego}& ICLR 2025& \textbf{61.3}& -&\\
 \midrule
Chat-UniVi \cite{jin2024chat}& CVPR 2024& 42.3& -& \multirow{2}{*}{OpenEQA}\\
 \cellcolor{gray!10}ALANAVLM \cite{suglia2024alanavlm}& EMNLP 2024& \textbf{46.7}& -& \\
 \midrule
  GPT-4o \cite{hurst2024gpt}&  arXiv 2024& -& \textbf{5.4}& \multirow{2}{*}{MULTIHOP-EGOQA}\\
\cellcolor{gray!10} GeLM \cite{chen2025grounded}& AAAI 2025& -& 4.8& \\
 \midrule
 EgoGPT (EgoIT) \cite{yang2025egolife}& CVPR 2025& 32.4& -& \multirow{2}{*}{EgoPlanbench}\\
\cellcolor{gray!10} EgoGPT (EgoIT+D1) \cite{yang2025egolife}& CVPR 2025& \textbf{33.4}& -& \\
 \midrule
 EgoGPT (EgoIT) \cite{yang2025egolife}& CVPR 2025& \textbf{61.7}& -& \multirow{3}{*}{EgoThink}\\
 \cellcolor{gray!10}EgoGPT (EgoIT+D1) \cite{yang2025egolife}& CVPR 2025& 61.4& -& \\
 RAVEN \cite{biswas2025raven}& arXiv 2025& 54.0& -& \\
 \midrule
\cellcolor{gray!10} EgoGPT (EgoIT) \cite{yang2025egolife}& CVPR 2025& 33.1& -& \multirow{2}{*}{EgolifeQA}\\
 EgoGPT (EgoIT+D1) \cite{yang2025egolife}& CVPR 2025& \textbf{36.0}& -& \\

 \bottomrule
    \end{tabular}
    
   \label{tab5}%
\end{table}

\subsection{Leveraging LLMs As Reasoning Engines}

These methods exploit pre-trained LLM reasoning without training large multimodal models, allowing efficient development. However, their accuracy is constrained by potential information loss during video-to-text conversion.

More suitable for solving egocentric long‑video question answering tasks, Wang et al. (Lifelongmemory) \cite{wang2023lifelongmemory} and Zhang et al. \cite{zhang2024hcqa,zhang2025hcqa} (HCQA/HCQA 1.5) are building on the powerful reasoning and descriptive capabilities of MLLMs, they generate textual descriptions of videos and leverage techniques such as context summarization and temporal localization, further incorporating in-context learning and reflection mechanisms, thereby achieving very high accuracy. Besides, Yang et al. \cite{yang2025egolife} (Egolife) even works on week-long videos. Unlike these methods, Taluzzi et al. \cite{taluzzi2025pixels} (FPG) construct graphs such as SceneNet and KnowledgeNet to obtain powerful reasoning and memory capabilities. It performs well on classification tasks and supports cross-modal reasoning, making it more suitable for relational analysis tasks.

\subsection{Training Specialized VideoQA Model}

In this paradigm, models are trained end-to-end on video data to integrate visual and linguistic features directly. This approach excels at capturing visual nuances and spatiotemporal relations but demands significant computational resources and large-scale annotated datasets.

Chen et al. (GeLM) \cite{chen2025grounded} propose an end-to-end multimodal large language model which employs grounded tokens ($<T>$, $</T>$) together with a dual-branch (saliency + similarity) evidence localization module. Its limitations lie in the reliance on manually curated multi-hop annotations and the insufficient handling of cross-modal ambiguities. With lower data annotation cost, Di et al. (GroundVQA) \cite{di2024grounded} achieves multimodal fusion with a Transformer, while a dual-head decoder predicts temporal segments and answers, supporting both open-ended and multiple-choice QA. Also, with low data requirements, Suglia et al. (ALANAVLM) \cite{suglia2024alanavlm} apply LoRA-based efficient fine-tuning to Chat-UniVi \cite{jin2024chat}, focusing on embedded scenario question answering. The approach yields a lightweight model (7B parameters) whose spatial reasoning performance surpasses that of GPT-4V. Fuse video, audio, and sensor modalities, Biswas et al. (RAVEN) \cite{biswas2025raven} dynamically filters irrelevant tokens via the QuART module. Combined with a three-stage training scheme, it exhibits strong cross-modal robustness. However, frame sampling in long videos often misses critical information. To better address long video processing, Ye et al. (MM-EGO) \cite{yemmego} rely on a memory-pointer prompting mechanism (global overview + key-frame fallback), achieving high efficiency in long-video processing and strong robustness against language bias. Huang et al. (Vinci) \cite{huang2025vinci} introduce a real-time intelligent assistant built on the EgoVideo-VL model, integrating a memory module and video generation and retrieval modules. It supports audio-based interaction and step-by-step visual demonstrations, offers strong real-time performance, and enables effective use of historical context and visually guided generation.
 
 \begin{table}
 \caption{Summarize the representative datasets in egocentric video question answering, Ego means egocentric view, and Exo means third-person view. Q-As represent question-answer pairs.}
     \centering
     \begin{tabular}{ccccccccc}
     \toprule
           \multicolumn{1}{c|}{\cellcolor[HTML]{EFEFEF}Dataset}&   \multicolumn{1}{c|}{\cellcolor[HTML]{EFEFEF}Years}&   \multicolumn{1}{c|}{\cellcolor[HTML]{EFEFEF}Duration}&   \multicolumn{1}{c|}{\cellcolor[HTML]{EFEFEF}Videos}&   \multicolumn{1}{c|}{\cellcolor[HTML]{EFEFEF}Segments}&    \multicolumn{1}{c|}{\cellcolor[HTML]{EFEFEF}Tasks}& \multicolumn{1}{c|}{\cellcolor[HTML]{EFEFEF}Q-As/Questions}&   \multicolumn{1}{c|}{\cellcolor[HTML]{EFEFEF}Domain}&  \multicolumn{1}{c}{\cellcolor[HTML]{EFEFEF}View}\\
          \midrule
           EQA \cite{das2018embodied}& CVPR 2018& -& -& -& 4& 5000+& Indoor&Ego\\
            AssistQ \cite{wong2022assistq}& ECCV 2022& 192h& 100& -& 25& 531& Household&Ego\\
          EgoSchema \cite{mangalam2023egoschema}&  NeurIPS 2023&  250h&  -&  844&   1&5063&  Mixed& Ego\\
           VIEW-QA \cite{song2024video}&  arXiv 2024&  10h&  1030&  -&   5&4120&  VIPs Assist& Ego\\
          EVUD \cite{suglia2024alanavlm}&  EMNLP 2024&  21.3h&  -&  2.4W&   9&10.5W&  Mixed& Ego\\
          HOI-QA \cite{bansal2024hoi}& arXiv 2024& -& 3956& -& 8& 390w& Mixed&Ego\\
          EgoTempo \cite{plizzari2025omnia}&  CVPR 2025&  274h&  221&  365&   10&500&  Mixed& Ego\\
          EgoBlind \cite{xiao2025egoblind}&  arXiv 2025&  9h&  478&  1329&   6&5311&  VIPs Assist& Ego\\
 EgoTextVQA \cite{zhou2025egotextvqa}& CVPR 2025& 253h& 1507& -& 11& 7064& Mixed&Ego\\ 
     \bottomrule
     \end{tabular}
     
    \label{tab6}%
 \end{table}

\subsection{Datasets}

Several relevant benchmarks and datasets have been introduced as shown in Table~\ref{tab6}.

\textbf{\textit{EQA}} \cite{das2018embodied} is a comprehensive benchmark for embodied agent training and evaluation, built on the House3D platform with over 45,000 SUNCG-based \cite{song2017semantic} 3D indoor scenes. It defines queries involving 12 room types and 50 object categories.
\textbf{\textit{AssistQ}} \cite{wong2022assistq} comprises 100 egocentric instructional videos (mean duration: 115 seconds) covering 25 household appliances, with 531 multi-step QA pairs. Questions are classified into general and specific functional categories, addressing appliance operation and settings.
\textbf{\textit{EgoSchema}} \cite{mangalam2023egoschema} comprises 5,063 video-question pairs drawn from 250 hours of Ego4D egocentric footage, with 3-minute clips annotated by $\geq$30 timestamps. It includes diverse question types, such as action reasoning and scene understanding. EgoSchema introduces the Temporal Certificate metric, which represents the minimal cumulative duration of segments required to verify annotation validity.
\textbf{\textit{VIEW-QA}} \cite{song2024video} comprises 1,030 videos (\~10 hours) and 4,120 QA pairs, with questions divided into five categories reflecting the practical needs of visually impaired users. Its dynamic, multi-task video QA design better mirrors real-world use cases.
\textbf{\textit{EVUD}} \cite{suglia2024alanavlm} supports both video captioning and QA tasks, comprising 13,849 Ego4D NLQ clips with 1,137 human-annotated QA pairs, 96,523 Gemini Pro 1.5-generated QA pairs across 7 categories, 7,680 polar QA pairs for spatial reasoning, 7,000 EgoClip videos with captions, and 3,475 short videos with captions generated via Habitat for HM3D scenes in OpenEQA \cite{majumdar2024openeqa}.
\textbf{\textit{HOI-QA}} \cite{bansal2024hoi} comprises 3.9 million QA pairs: 1.8M from EPIC-Kitchens (with narratives, object bounding boxes, and hand-object contact) and 2.1M from Ego4D’s FHO benchmark (narratives, hand/object bounding boxes). The data is split into 3.2M training samples and 0.75M test samples.
\textbf{\textit{EgoTempo}} \cite{plizzari2025omnia} features 500 QA pairs spanning 10 temporal reasoning tasks using egocentric videos (avg. 45 seconds) from 40 scenarios, designed to evaluate MLLM temporal integration abilities.
\textbf{\textit{EgoBlind}} \cite{xiao2025egoblind} is the first QA dataset constructed from real-world egocentric videos recorded by blind individuals. Unlike prior datasets such as VizWiz (static images) \cite{gurari2018vizwiz} and VIEW-QA (simulated blindness) \cite{song2024video}, EgoBlind uniquely addresses the dynamic informational needs of blind users, which are not captured by general video QA resources.
\textbf{\textit{EgoTextVQA}} \cite{zhou2025egotextvqa} targets scene text understanding in egocentric contexts, comprising 1,507 curated indoor and outdoor videos and 7,064 QA pairs. Videos were filtered using OCR \cite{li2022pp} from RoadTextVQA \cite{tom2023reading}, EgoSchema, and Ego4D, with manual quality control to ensure the presence of relevant text.

\subsection{Evaluation}

Evaluating egocentric video question answering requires a comprehensive assessment of memory, reasoning, comprehension, and interpretability. Therefore, certain benchmarks \cite{mangalam2023egoschema}\cite{biswas2025raven}\cite{yang2025egolife} are typically selected for testing. 

In benchmark evaluations, accuracy \cite{yang2025egolife} is frequently employed, defined in multiple-choice question answering as the proportion of correctly answered questions to the total.

Some approaches employ "Sent.Sim." \cite{chen2025grounded}, which uses Sentence-BERT cosine similarity to assess the semantic alignment between predicted and ground truth answers. "Comp." evaluates whether the answer comprehensively incorporates all multimodal evidence required by the question. "Coher." \cite{biswas2025raven} assesses logical and semantic coherence, ensuring that answers are both contradiction-free and contextually consistent.

\begin{figure}
    \centering
    \includegraphics[width=1\linewidth]{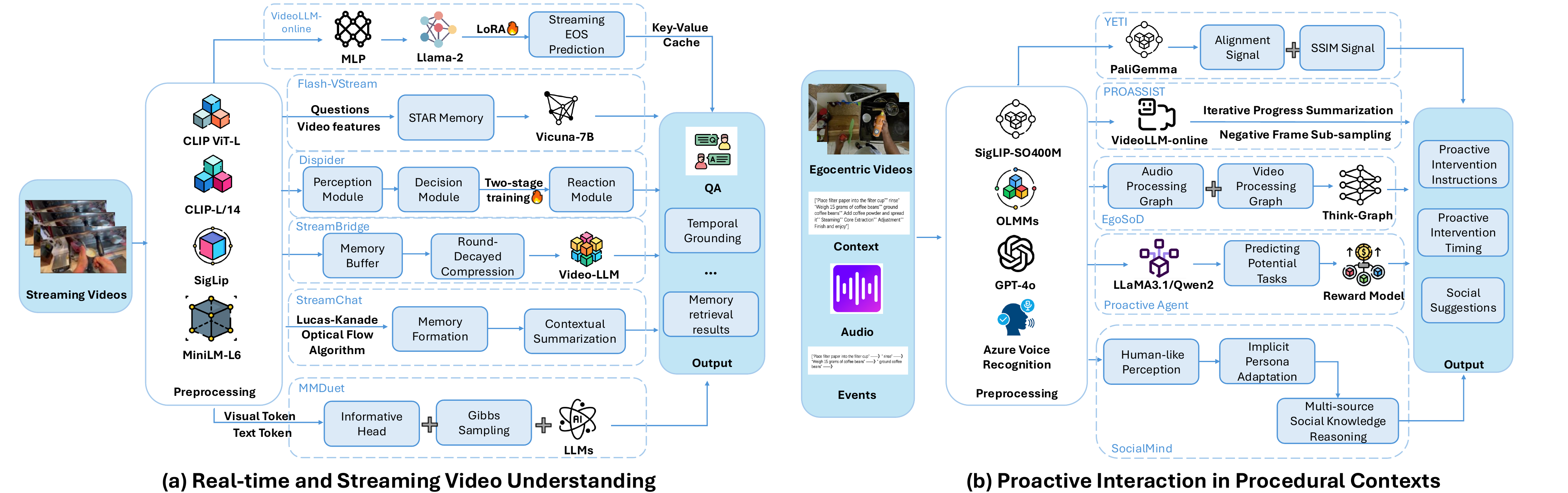}
  \caption{(a) shows the different methods in real-time and streaming video understanding: VideoLLM-online \cite{chen2024videollm}, Flash-VStream \cite{zhang2024flash}, Dispider \cite{qian2025dispider}, StreamBridge \cite{wang2025streambridge}, StreamChat \cite{xiongstreaming}, MMDuet \cite{wang2024videollm}, and (b) shows that in proactive interaction in procedural contexts: YETI \cite{bandyopadhyay2025yeti}, PROASSIST \cite{zhang2025proactive}, EgoSoD \cite{wang2025egosocial}, Proactive Agent \cite{luproactive}, SocialMind \cite{yang2025socialmind}.}
    \label{fig8}%
\end{figure}

\section{Enabling Dimensions for Real-World EgoProceAssist}

In this section, we formalize two enabling dimensions and introduce their definitions, key methods, datasets, and evaluation protocols, respectively.

\subsection{Real-time and Streaming Video Understanding}

\subsubsection{Definition}

Real-time and streaming video understanding is an intelligent analysis paradigm for dynamic, continuous video streams, thereby enabling perception and interpretation of ongoing visual content. By moving beyond the traditional offline setting of processing complete videos in a single pass and instead aligning with the generate-and-process simultaneously nature of real-world video, it has become a key enabling technology for applications such as autonomous driving \cite{zhong2025cooptrack}, real-time surveillance, and robotic interaction \cite{yao2025inference}.

Rising demand for online, real-time video processing has focused attention on real-time and streaming video understanding and related tasks \cite{de2016online,zhao2022real,wang2021oadtr}, such as real-time hand gesture recognition \cite{benitez2021ipn} and 3D pose estimation \cite{luvizon2020multi}. Recent multimodal models further advance real-time interaction across vision, audio, and text \cite{fu2024vita,fu2025vita,xie2024mini}. Aligned with our research, many existing methods already support real-time responses. For example, MiniROAD in PREGO \cite{flaborea2024prego} and TI-PREGO \cite{plini2025ti} are inherently real-time action-detection branches. Likewise, GC \cite{mazzamuto2025gazing} uses gaze prediction to detect errors. Although not always stated, gaze signals are intrinsically streaming and have been experimentally used for online intervention.

\begin{table}
\caption{Performance comparison of representative methods for real-time and streaming video understanding, where metric values are averaged across different tasks within the same dataset. The bold text represents the best performance on this metric. Score (0-5) represents the LLM Judge's results.}
    \centering
    \begin{tabular}{cccccc|c|}
    \toprule
         \multicolumn{1}{c|}{\cellcolor[HTML]{EFEFEF}Method}& \multicolumn{1}{c|}{\cellcolor[HTML]{EFEFEF}Year} &  \multicolumn{1}{c|}{\cellcolor[HTML]{EFEFEF}Accuracy}&  \multicolumn{1}{c|}{\cellcolor[HTML]{EFEFEF}Score (0-5)}&  \multicolumn{1}{c|}{\cellcolor[HTML]{EFEFEF}Coh.} &\multicolumn{1}{c|}{\cellcolor[HTML]{EFEFEF}F1}&  \multicolumn{1}{c|}{\cellcolor[HTML]{EFEFEF}Benchmark}\\
         \midrule
         
         VideoLLM-online \cite{chen2024videollm}&  CVPR 2024&  -&  -&   - &13.7&  \multirow{2}{*}{ETBench}\\
         \cellcolor{gray!10}Dispider \cite{qian2025dispider}&  CVPR 2025&  -&  -&  -&  \textbf{31.0}&  \\
         \midrule
         \
         VideoLLM-online \cite{chen2024videollm}&  CVPR 2023&  36.0&  -&  -&  -&\multirow{10}{*}{StreamingBench}  \\

         \cellcolor{gray!10}Claude-3.5-Sonnet \cite{bae2024enhancing}& -& 72.4& -& - &-&\\
 GPT-4o \cite{hurst2024gpt}& arXiv 2024& 77.1& -& - &-& \\
 \cellcolor{gray!10}Gemini-1.5-Pro \cite{team2024gemini}&  arXiv 2024&  \textbf{79.0}&  -&   - &-&  \\

 Flash-VStream \cite{zhang2024flash}& arXiv 2024& 23.2& -& -& -& \\
 \cellcolor{gray!10}IXC2.5-OL \cite{zhang2024internlm}& arXiv 2024& 73.8& -& -& -& \\
  Dispider \cite{qian2025dispider}& CVPR 2025& 67.6& -& -& -& \\
 \cellcolor{gray!10}Oryx-1.5-7B (StreamBridge) \cite{wang2025streambridge}& arXiv 2025& 74.8& -& -& -& \\
 LLaVA-OV-7B (StreamBridge) \cite{wang2025streambridge}& arXiv 2025& 70.9&  -& - &-&\\
 \cellcolor{gray!10}Qwen2-VL-7B (StreamBridge) \cite{wang2025streambridge}& arXiv 2025& 77.0& -&  - &-& \\

 \midrule

 Flash-VStream \cite{zhang2024flash}& arXiv 2024& 33.6& -& -& -& \multirow{5}{*}{OvO-Bench}\\
 \cellcolor{gray!10}Dispider \cite{qian2025dispider}&  CVPR 2025&  41.8&  -&  -&  - &  \\
 Oryx-1.5-7B (StreamBridge) \cite{wang2025streambridge}& arXiv 2025& 71.2& -& -& -& \\
 \cellcolor{gray!10}LLaVA-OV-7B (StreamBridge) \cite{wang2025streambridge}& arXiv 2025& 69.9& -& - &-&\\
 Qwen2-VL-7B (StreamBridge) \cite{wang2025streambridge}& arXiv 2025& \textbf{71.3}& -& - &-& \\
 \midrule
 
 \cellcolor{gray!10} GPT-4o-mini \cite{hurst2024gpt}&  arXiv 2024& 59.1& 3.2& 2.0& -& \multirow{4}{*}{StreamBench}\\
 VideoLLM-online (FPS: 5) \cite{chen2024videollm}& CVPR 2024& 56.4& 3.1& 1.9&-& \\

 \cellcolor{gray!10}Flash-VStream (FPS: 1) \cite{zhang2024flash}& arXiv 2024& 52.1& 2.9& \textbf{2.2}& -& \\

  StreamChat (FPS: 20) \cite{xiongstreaming}& ICLR 2025& \textbf{63.8}& \textbf{3.4}& 1.79& -& \\

 \bottomrule
    \end{tabular}
    
   \label{tab7}%
\end{table}

\subsubsection{Methods}

As shown in Fig.~\ref{fig8} and Table~\ref{tab7}, VideoLLM-online \cite{chen2024videollm} uses streaming EOS prediction to suppress redundant responses and improve inference efficiency, but its single-LLM architecture cannot parallelize perception and reasoning, leading to blocked frame processing during long responses. To address this issue, Dispider \cite{qian2025dispider} decomposes the system into three asynchronous modules, enabling parallel perception and reasoning and alleviating blocking, but at the cost of higher system complexity, greater deployment overhead, and weaker support for short-video offline tasks compared with VideoLLM-online. IXC2.5-OL \cite{zhang2024internlm} performs multimodal streaming over audio and video, extending its applicability to more complex scenarios. However, multimodal processing increases computational and resource costs, and its global-matching memory retrieval performs poorly on fine-grained temporal localization. Flash-VStream \cite{zhang2024flash} addresses this via the STAR memory mechanism, which partitions memory into four complementary sub-modules to compress information and enhance retrieval efficiency. StreamBridge \cite{wang2025streambridge} is a training-free framework that can be directly applied to offline Video-LLMs, using a Round-Decayed Compression strategy to dynamically compress historical content while emphasizing recent information. StreamChat \cite{xiongstreaming} is also training-free and organizes short-term, long-term, and dialogue memory hierarchically to support parallel processing, but its fixed frame-selection threshold risks information loss in highly dynamic scenes. VideoStreaming \cite{qian2024streaming} shares similar limitations. It maintains a constant token length for videos of arbitrary duration, stabilizing memory and inference costs, but cannot adapt the representation length to content complexity, which can lead to redundancy or information loss. MMDuet \cite{wang2024videollm} evaluates each frame’s information value and query relevance in real time to decide whether to interrupt the stream and respond, but it exhibits slow inference and a tendency toward repetitive outputs. To solve it, Video-MoD \cite{wu2024videollm} inserts a LayerExpert module into selected Transformer layers to dynamically select key visual tokens per frame (typically retaining 20\%), applying self-attention and feed-forward operations only to these tokens. This reduces redundancy, lowers memory requirements, and improves efficiency.
\begin{table}
\caption{Summarize the representative datasets and benchmarks in real-time and streaming video understanding, Ego means egocentric view and Exo means third-person view.}
    \centering

     \begin{tabular}{cccccccc}
     \toprule
        \multicolumn{1}{c|}{\cellcolor[HTML]{EFEFEF}Dataset}&   \multicolumn{1}{c|}{ \cellcolor[HTML]{EFEFEF}Years}&   \multicolumn{1}{c|}{ \cellcolor[HTML]{EFEFEF}Duration}&   \multicolumn{1}{c|}{\cellcolor[HTML]{EFEFEF}Videos}&   \multicolumn{1}{c|}{ \cellcolor[HTML]{EFEFEF}QA-pairs}&   \multicolumn{1}{c|}{ \cellcolor[HTML]{EFEFEF}tasks}&   \multicolumn{1}{c|}{ \cellcolor[HTML]{EFEFEF}Domain}&  \multicolumn{1}{c}{ \cellcolor[HTML]{EFEFEF}View} \\
         \midrule
         MMDuetIT \cite{wang2024videollm}&  arXiv 2024&  -&  -&  -&  3&Mixed&Ego+Exo \\
          StreamingBench \cite{lin2024streamingbench} &  arXiv 2024&  -&  900&  4300&  18&Mixed& Ego+Exo \\
          OVO-Bench \cite{niu2025ovo}& CVPR 2025& 75h& 644& 2814& 12&Mixed&Ego+Exo \\
          SVBench \cite{yangsvbench}&  ICLR 2025& 56.4h& 1353& 49979& 9&Mixed&Ego+Exo \\       
         Stream-IT \cite{wang2025streambridge} &  arXiv 2025&  -& -& 120k&  8&Mixed& Ego+Exo \\
   \bottomrule
      \end{tabular}

\label{tab8}%
\end{table}

\subsubsection{Datasets}

Several related datasets and benchmarks merit particular attention as shown in Table~\ref{tab8}.
\textbf{\textit{MMDuetIT}} \cite{wang2024videollm} comprises approximately 109,000 training samples spanning multiple video sources, including Shot2Story \cite{han2023shot2story20k}, COIN \cite{tang2019coin}, DiDeMo \cite{anne2017localizing}, and is further enhanced for real-time streaming video understanding by multi-answer grounded video QA data (Shot2Story-MAGQA-39k) generated with GPT-4o.
\textbf{\textit{StreamingBench}} \cite{lin2024streamingbench} formalizes three core dimensions of streaming video comprehension: real-time visual understanding, holistic source understanding, and contextual understanding, and provides a dataset of 18 tasks, 900 videos, and 4,300 human-annotated QA pairs, spanning eight real-world scenarios such as lifelogging, competitions, education, and more.
\textbf{\textit{OVO-Bench}} \cite{niu2025ovo} systematically evaluates the temporal perception and dynamic reasoning capabilities of VLMs from three key dimensions: backward tracing, real-time visual perception, and forward active responding. It includes 644 unique videos and constructs approximately 2,814 finely grained meta-annotated samples with precise timestamps, covering 12 specific tasks.
\textbf{\textit{SVBench}} \cite{yangsvbench} consists of 1,353 diverse videos sourced from 6 streaming platforms, with an average video length of more than 2 minutes, generating 49,979 high-quality QA-pairs, with an average of 36.94 pairs per video, which is the highest average number of QA-pairs among currently known datasets.
\textbf{\textit{Stream-IT}} \cite{wang2025streambridge} integrates multiple publicly available video annotation datasets (e.g., ActivityNet \cite{caba2015activitynet}, Shot2Story \cite{han2023shot2story20k}, COIN \cite{tang2019coin}) and incorporates the large-scale synthetic dataset StreamingQA-120K \cite{wang2025streambridge}, yielding approximately 1.8 million training samples that span dense video captioning, procedural step recognition, grounded video question answering, and temporal video localization.

\subsubsection{Evaluation}

Existing works typically evaluate real-time and streaming video understanding via multi-perspective video question answering (VQA), temporal grounding, and event localization. Some also measure models’ processing speed and resource usage on video streams. For question answering or decision-making tasks, accuracy is the dominant metric. Increasingly, the LLM-as-a-Judge paradigm is adopted, using LLMs to score responses. F1, mIoP, mIoU, and TimeDiff assess temporal grounding and event query performance, while Coh. evaluates conversational continuity in multi-turn dialogue. FPS and VRAM are used to quantify inference speed and resource consumption.

\subsection{Proactive Interaction in Procedural Contexts}

\subsubsection{Definition}

Proactive interaction is a mode in which an intelligent agent (e.g., a dialogue \cite{zhang2025proactive,deng2023survey} or multimodal assistant \cite{berube2024proactive}) autonomously initiates actions, provides information, or adapts interaction strategies, without explicit user commands, based on continuous perception of context, user state, and task goals, to satisfy explicit or latent user needs. Its timing, content, and form must match the scenario and user expectations to avoid intrusiveness.

With the growing deployment of user-facing assistants \cite{luproactive}, research on proactive interaction has emerged across multiple domains \cite{wang2025egosocial}. Proactivity has been extensively studied in social intent analysis \cite{wang2025egosocial,yang2025socialmind,zhang2024proagent} and human–computer interaction \cite{zhang2024proagent} and is gaining attention in work on speech-based voice assistants \cite{berube2024proactive}. Within the three core functions of EgoProceAssist, several methods already support proactive assistance: PREGO \cite{flaborea2024prego} and TI-PREGO \cite{plini2025ti} use predicted future action sets for error detection and can proactively warn users of errors; VINCI \cite{huang2025vinci} performs proactive task planning by suggesting next steps; and HoloAssist dataset \cite{wang2023holoassist} provides explicit labels for proactive interventions. These capabilities make such approaches particularly important for future developments in this field.

\subsubsection{Methods}

Existing methods for proactive interaction cover many application domains, yet specialized approaches for egocentric vision remain relatively limited. As shown in Fig.~\ref{fig8}, SocialMind \cite{yang2025socialmind} captures verbal, non-verbal, and social cues and employs LLM-based reasoning to generate proactive, real-time social suggestions. Tailored to AR hardware, it integrates multimodal signals widely used in first-person vision, thereby aligning closely with egocentric vision research. However, local processing of rich multimodal streams introduces substantial hardware overhead and heightens privacy risks. To reduce this burden, YETI \cite{bandyopadhyay2025yeti} adopts a lightweight, signal-driven paradigm, using video frame structural similarity (SSIM) and object count variation (Alignment Signal) to rapidly decide when to intervene. Also based on LLMs' reasoning capabilities, ProAgent \cite{zhang2024proagent} proposes a modular Plan-Verify-Control-Memory framework, thereby improving both human-agent collaboration and coordination among multiple agents. Its reliance on predefined task rules and skill libraries, however, restricts generalization to open-ended scenarios, and its memory design offers only limited optimization for long-horizon collaboration. Proactive Agent \cite{luproactive} addresses these memory constraints via a hierarchical architecture that decouples long-term knowledge storage from short-term trajectory memory. EgoSoD \cite{wang2025egosocial} advances proactive interaction by constructing a social cognition graph that integrates audio and visual cues, and by determining intervention timing through a proximity–gaze–speech hierarchical filtering mechanism. PROASSIST \cite{zhang2025proactive} further tackles the severe imbalance between response and non-response frames using Negative Frame Sub-sampling (NFS), and achieves efficient long-video handling through Iterative Progress Summarization (IPS) for dynamic temporal compression and summarization.

\begin{table}
\caption{Summarize the representative datasets and benchmarks in proactive interaction, Ego means egocentric view and Exo means third-person view.}
    \centering

     \begin{tabular}{cccccccc}
     \toprule
        \multicolumn{1}{c|}{\cellcolor[HTML]{EFEFEF}Dataset}&   \multicolumn{1}{c|}{ \cellcolor[HTML]{EFEFEF}Years}&   \multicolumn{1}{c|}{ \cellcolor[HTML]{EFEFEF}Duration}&   \multicolumn{1}{c|}{\cellcolor[HTML]{EFEFEF}Videos}&   \multicolumn{1}{c|}{ \cellcolor[HTML]{EFEFEF}QA-pairs}&   \multicolumn{1}{c|}{ \cellcolor[HTML]{EFEFEF}tasks}&   \multicolumn{1}{c|}{ \cellcolor[HTML]{EFEFEF}Domain}&  \multicolumn{1}{c}{ \cellcolor[HTML]{EFEFEF}View} \\
         \midrule
          ProactiveBench \cite{luproactive} &  ICLR 2025& -& -& -& 12&Mixed&- \\       
         ProactiveVideoQA \cite{wang2025proactivevideoqa} &  arXiv 2025&  47.7h&  1377&  1427&  4&Mixed&Ego+Exo \\
          STREAMGAZE \cite{lee2025streamgaze} &  arXiv 2025&  66.2h&  285&  8521&  10&Mixed& Ego \\
          PROASSIST \cite{zhang2025proactive} & arXiv 2025& 479h& 3934& 30135& 4&Mixed&Ego \\
         
         EgoSocial \cite{wang2025egosocial} &  arXiv 2025&  4.2h&-& 13500&  -&Mixed& Ego \\
   \bottomrule
      \end{tabular}

\label{tab9}%
\end{table}

\subsubsection{Datasets}

Besides HoloAssist mentioned above, there are several other important datasets and benchmarks that deserve attention, as shown in Table~\ref{tab9}.

\textbf{\textit{ProactiveBench}} \cite{luproactive} includes 6,790 training events and 233 test events across coding, writing, and daily life, offering large-scale, realistic contexts for training and evaluating the proactive behavior of large language model agents. The authors also built a reward-model subset of 1,760 annotated instances to emulate human acceptance or rejection of task predictions.
\textbf{\textit{ProactiveVideoQA}} \cite{wang2025proactivevideoqa} is the first benchmark focused on evaluating proactive interaction. It combines four tasks: web video QA, egocentric video QA, TV series QA, and video anomaly detection, includes 1,377 videos and 1,427 QA pairs, with about 3.5 turns per video.
\textbf{\textit{STREAMGAZE}} \cite{lee2025streamgaze} defines 10 tasks spanning past, present, and future (including object recognition, gaze-sequence matching, and future action prediction) to comprehensively assess models’ understanding of gaze dynamics and intention prediction, contains 285 videos and 8,521 QA pairs covering real-world interactive scenarios.
\textbf{\textit{PROASSIST}} \cite{zhang2025proactive} constructs dialogue data from the temporal annotations of six public egocentric video datasets, covering a variety of task scenarios such as cooking, object manipulation, and experiments, comprises 30,135 dialogues spanning 479 hours of video, with an average video length of approximately 16 minutes, and provides proactivity and intent labels as well as progress summaries for each assistant turn.
\textbf{\textit{EgoSocial}} \cite{wang2025egosocial} focuses on egocentric social interaction understanding and evaluates multimodal large models’ proactive intervention capabilities in social scenarios. It contains 13,500 video–question pairs annotated with eight categories of social cues, such as speech activity, gaze, and personal space, and it establishes the first quantitative evaluation protocol for social intervention detection.

\subsubsection{Evaluation}

Proactive model performance is typically evaluated using precision, recall, F1-score, and accuracy to measure the correctness and completeness of predictions. Many studies also tailor metrics to task characteristics. For example, Proactive Agent \cite{luproactive} adds false-alarm rate to capture unnecessary interventions, while EgoSocial \cite{wang2025egosocial} uses intervention timing (accuracy) and macro-F1 to assess precise, non-disruptive interventions across diverse social scenarios. Some approaches further introduce specialized indicators, such as acceptance rate in Proactive Agent \cite{luproactive}, or using the LLM-as-a-Judge strategy. GPT-4o, for instance, provides five-point ratings for the correctness, timeliness, and efficiency of proactive behaviors. Human evaluations additionally cover subjective aspects such as conversational naturalness and task alignment, as in SocialMind \cite{yang2025socialmind} and PROASSIST \cite{zhang2025proactive}.

Because these methods are evaluated on distinct, domain-specific datasets, their quantitative results are not directly comparable. Accordingly, this section does not include a unified tabular comparison of performance metrics.

\begin{figure}
    \centering
    \includegraphics[width=1\linewidth]{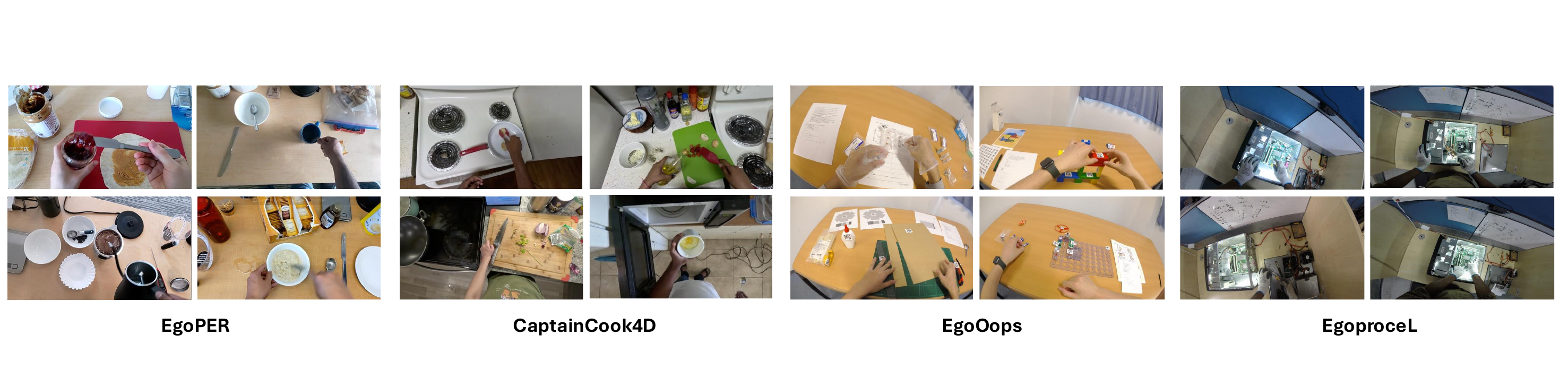}
    \caption{Examples of datasets, EgoPER \cite{lee2024error} and CaptainCook4D \cite{peddi2024captaincook4d} focus on kitchen cooking scenarios, EgoOops \cite{haneji2025egooops} designs complex scenarios such as chemical experiments and building blocks, and EgoproceL \cite{bansal2022my} focuses on pc assembly and disassembly tasks.}
    \label{fig9}%
\end{figure}

\section{Experiments}

This section presents supplementary experiments, designed to show that popular vision-language models still have substantial room for improvement in assisting with procedural tasks. We begin by introducing the four selected datasets shown in Fig.~\ref{fig9}, followed by a detailed description of the experimental procedures. Finally, we present a comparative analysis of the experimental results using tables and provide an in-depth discussion. More detailed experimental information and visual results are provided in the attached \textcolor{blue}{Appendix} file.

\begin{table}
\caption{Experimental results in egocentric procedural error detection, the part highlighted in purple is the result of our supplementary experiments, bold text represents the best performance on this metric in each part, and Time is the average inference time. ICL denotes results under an in-context learning setting, and the green upward arrow indicates performance improvement.}
    \centering
    \begin{tabular}{ccccccccc|c|}
    \toprule
         \multicolumn{1}{c|}{ \cellcolor[HTML]{EFEFEF}Method}&   \multicolumn{1}{c|}{ \cellcolor[HTML]{EFEFEF}Years} &\multicolumn{1}{c|}{ \cellcolor[HTML]{EFEFEF}Time}&   \multicolumn{1}{c|}{\cellcolor[HTML]{EFEFEF}Precision}&   \multicolumn{1}{c|}{ \cellcolor[HTML]{EFEFEF}EDA}&   \multicolumn{1}{c|}{ \cellcolor[HTML]{EFEFEF}AUC}&   \multicolumn{1}{c|}{ \cellcolor[HTML]{EFEFEF}F1}&   \multicolumn{1}{c|}{ \cellcolor[HTML]{EFEFEF}Recall}&   \multicolumn{1}{c|}{ \cellcolor[HTML]{EFEFEF}Acc}&  \multicolumn{1}{c|}{ \cellcolor[HTML]{EFEFEF}Dataset}\\
         \midrule
         EgoPED \cite{lee2024error}& CVPR 2024 &-& 56.5& 69.8& 54.9& -& -& 69.8&\multirow{10}{*}{CaptainCook4D}\\
 AMNAR \cite{huang2025modeling}& CVPR 2025 &-& \textbf{66.8}& \textbf{72.3}& \textbf{60.2}& -& -& \textbf{72.3}&\\
\cellcolor{blue!7}Just Ask \cite{yang2021just}&  \cellcolor{blue!7}ICCV 2021 &\cellcolor{blue!7}-& \cellcolor{blue!7}35.1&  \cellcolor{blue!7}35.1&  \cellcolor{blue!7}50.0&  \cellcolor{blue!7}\textbf{51.2}&  \cellcolor{blue!7}-&  \cellcolor{blue!7}35.1& \\
  
\cellcolor{blue!7}Video-LLaVA \cite{lin2024video}&  \cellcolor{blue!7}EMNLP 2024 &\cellcolor{blue!7}\textbf{2.55}& \cellcolor{blue!7}\textbf{40.1}&  \cellcolor{blue!7}60.8&  \cellcolor{blue!7}52.2&  \cellcolor{blue!7}29.6&  \cellcolor{blue!7}23.4&  \cellcolor{blue!7}60.8& \\

\cellcolor{blue!7}Video-LLaMA2 \cite{cheng2024videollama}&  \cellcolor{blue!7}arXiv 2024 &\cellcolor{blue!7}4.73& \cellcolor{blue!7}36.5&  \cellcolor{blue!7}58.4&  \cellcolor{blue!7}50.1&  \cellcolor{blue!7}29.8&  \cellcolor{blue!7}25.2&  \cellcolor{blue!7}58.4& \\
\cellcolor{blue!7}{Vinci} \cite{huang2025vinci}&  \cellcolor{blue!7}{IMWUT 2025} &\cellcolor{blue!7}{3.03}&  \cellcolor{blue!7}{35.6}&  \cellcolor{blue!7}{49.7}&  \cellcolor{blue!7}{48.5}&  \cellcolor{blue!7}{-}&  \cellcolor{blue!7}{-}&  \cellcolor{blue!7}{49.7}&\\
\cellcolor{blue!7}LLaVA-OneVision \cite{lillava}&  \cellcolor{blue!7}TMLR 2025 &\cellcolor{blue!7}4.55& \cellcolor{blue!7}39.8&  \cellcolor{blue!7}\textbf{63.0}&  \cellcolor{blue!7}\textbf{53.1}&  \cellcolor{blue!7}17.0&  \cellcolor{blue!7}10.8&  \cellcolor{blue!7}\textbf{63.0}& \\
 \cellcolor{blue!7}{EgoGPT} \cite{yang2025egolife}&  \cellcolor{blue!7}{CVPR 2025} &\cellcolor{blue!7}{3.82}&  \cellcolor{blue!7}{36.0}&  \cellcolor{blue!7}{46.5}&  \cellcolor{blue!7}{49.6}&  \cellcolor{blue!7}{47.0}&  \cellcolor{blue!7}{\textbf{67.5}}&  \cellcolor{blue!7}{46.5}&\\
 \cellcolor{blue!7}{Qwen2.5-Omni-7B} \cite{xu2025qwen2}&  \cellcolor{blue!7}{arXiv 2025} &\cellcolor{blue!7}{4.19}&  \cellcolor{blue!7}{32.3}&  \cellcolor{blue!7}{59.7}&  \cellcolor{blue!7}{53.0}&  \cellcolor{blue!7}{19.1}&  \cellcolor{blue!7}{13.6}&  \cellcolor{blue!7}{59.7}&\\
 \cellcolor{blue!7}{Ming-Lite-Omni} \cite{ai2025ming}&  \cellcolor{blue!7}{arXiv 2025} &\cellcolor{blue!7}{11.5}&  \cellcolor{blue!7}{32.6}&  \cellcolor{blue!7}{51.4}&  \cellcolor{blue!7}{45.7}&  \cellcolor{blue!7}{34.2}&  \cellcolor{blue!7}{36.0}&  \cellcolor{blue!7}{51.4}&\\
   \midrule

\cellcolor{blue!7}Video-LLaVA \cite{lin2024video}&  \cellcolor{blue!7}EMNLP 2024 &\cellcolor{blue!7}\textbf{1.71}& \cellcolor{blue!7}14.6&  \cellcolor{blue!7}60.0&  \cellcolor{blue!7}44.6&  \cellcolor{blue!7}15.8&  \cellcolor{blue!7}17.1&  \cellcolor{blue!7}60.0& \multirow{14}{*}{EgoOops}\\
\cellcolor{blue!15}{{Video-LLaVA (ICL)}} \cite{lin2024video}&  \cellcolor{blue!15}{EMNLP 2024} &\cellcolor{blue!15}{1.94}&  \cellcolor{blue!15}{20.9 \textcolor{green}{$\uparrow$}}&  \cellcolor{blue!15}{62.5 \textcolor{green}{$\uparrow$}}&  \cellcolor{blue!15}{49.3 \textcolor{green}{$\uparrow$}}&  \cellcolor{blue!15}{23.1 \textcolor{green}{$\uparrow$}}&  \cellcolor{blue!15}{25.7 \textcolor{green}{$\uparrow$}}&  \cellcolor{blue!15}{62.5 \textcolor{green}{$\uparrow$}}&\\
\cellcolor{blue!7}Video-LLaMA2 \cite{cheng2024videollama}&  \cellcolor{blue!7}arXiv 2024 &\cellcolor{blue!7}3.51& \cellcolor{blue!7}15.4&  \cellcolor{blue!7}50.0&  \cellcolor{blue!7}42.4&  \cellcolor{blue!7}20.0&  \cellcolor{blue!7}28.6&  \cellcolor{blue!7}50.0& \\
\cellcolor{blue!15}{{Video-LLaMA2 (ICL)}} \cite{cheng2024videollama}&  \cellcolor{blue!15}{arXiv 2024} &\cellcolor{blue!15}{3.59}&  \cellcolor{blue!15}{29.1 \textcolor{green}{$\uparrow$}}&  \cellcolor{blue!15}{57.5 \textcolor{green}{$\uparrow$}}&  \cellcolor{blue!15}{59.4 \textcolor{green}{$\uparrow$}}&  \cellcolor{blue!15}{40.4 \textcolor{green}{$\uparrow$}}&  \cellcolor{blue!15}{65.7 \textcolor{green}{$\uparrow$}}&  \cellcolor{blue!15}{57.5 \textcolor{green}{$\uparrow$}}&\\
\cellcolor{blue!7}{Vinci} \cite{huang2025vinci}&  \cellcolor{blue!7}{IMWUT 2025} &\cellcolor{blue!7}{4.39}&  \cellcolor{blue!7}{19.4}&  \cellcolor{blue!7}{26.9}&  \cellcolor{blue!7}{47.3}&  \cellcolor{blue!7}{30.8}&  \cellcolor{blue!7}{74.3}&  \cellcolor{blue!7}{26.9}&\\
\cellcolor{blue!15}{{Vinci (ICL)}} \cite{huang2025vinci}&  \cellcolor{blue!15}{IMWUT 2025} &\cellcolor{blue!15}{4.46}&  \cellcolor{blue!15}{25.5 \textcolor{green}{$\uparrow$}}&  \cellcolor{blue!15}{49.4 \textcolor{green}{$\uparrow$}}&  \cellcolor{blue!15}{58.9 \textcolor{green}{$\uparrow$}}&  \cellcolor{blue!15}{37.2 \textcolor{green}{$\uparrow$}}&  \cellcolor{blue!15}{68.6}&  \cellcolor{blue!15}{49.4 \textcolor{green}{$\uparrow$}}&\\
\cellcolor{blue!7}LLaVA-OneVision \cite{lillava}&  \cellcolor{blue!7}TMLR 2025 &\cellcolor{blue!7}1.91& \cellcolor{blue!7}26.1&  \cellcolor{blue!7}64.4&  \cellcolor{blue!7}54.1&  \cellcolor{blue!7}29.6&  \cellcolor{blue!7}34.3&  \cellcolor{blue!7}64.4& \\
\cellcolor{blue!15}{{LLaVA-OneVision (ICL)}} \cite{lillava}&  \cellcolor{blue!15}{TMLR 2025} &\cellcolor{blue!15}{1.89}&  \cellcolor{blue!15}{29.6 \textcolor{green}{$\uparrow$}}&  \cellcolor{blue!15}{71.3 \textcolor{green}{$\uparrow$}}&  \cellcolor{blue!15}{52.6}&  \cellcolor{blue!15}{25.8}&  \cellcolor{blue!15}{22.9}&  \cellcolor{blue!15}{71.3 \textcolor{green}{$\uparrow$}}&\\
\cellcolor{blue!7}{EgoGPT} \cite{yang2025egolife}&  \cellcolor{blue!7}{CVPR 2025} &\cellcolor{blue!7}{3.27}&  \cellcolor{blue!7}{\textbf{34.8}}&  \cellcolor{blue!7}{\textbf{73.8}}&  \cellcolor{blue!7}{55.1}&  \cellcolor{blue!7}{27.6}&  \cellcolor{blue!7}{22.9}&  \cellcolor{blue!7}{\textbf{73.8}}&\\
\cellcolor{blue!15}{{EgoGPT (ICL)}} \cite{yang2025egolife}&  \cellcolor{blue!15}{CVPR 2025} &\cellcolor{blue!15}{3.38}&  \cellcolor{blue!15}{30.8}&  \cellcolor{blue!15}{68.8}&  \cellcolor{blue!15}{53.6}&  \cellcolor{blue!15}{32.4 \textcolor{green}{$\uparrow$}}&  \cellcolor{blue!15}{34.3 \textcolor{green}{$\uparrow$}}&  \cellcolor{blue!15}{68.8}&\\
\cellcolor{blue!7}{Qwen2.5-Omni-7B} \cite{xu2025qwen2}&  \cellcolor{blue!7}{arXiv 2025} &\cellcolor{blue!7}{6.48} &  \cellcolor{blue!7}{10.6}&  \cellcolor{blue!7}{45.6}&  \cellcolor{blue!7}{36.5}&  \cellcolor{blue!7}{13.9}&  \cellcolor{blue!7}{20.0}&  \cellcolor{blue!7}{45.6}&\\
\cellcolor{blue!15}{{Qwen2.5-Omni-7B (ICL)}} \cite{xu2025qwen2}&  \cellcolor{blue!15}{arXiv 2025} &\cellcolor{blue!15}{6.50 \textcolor{green}{$\uparrow$}}&  \cellcolor{blue!15}{26.7 \textcolor{green}{$\uparrow$}}&  \cellcolor{blue!15}{48.8 \textcolor{green}{$\uparrow$}}&  \cellcolor{blue!15}{63.4 \textcolor{green}{$\uparrow$}}&  \cellcolor{blue!15}{39.7 \textcolor{green}{$\uparrow$}}&  \cellcolor{blue!15}{77.1 \textcolor{green}{$\uparrow$}}&  \cellcolor{blue!15}{48.8 \textcolor{green}{$\uparrow$}}&\\
\cellcolor{blue!7}{Ming-Lite-Omni} \cite{ai2025ming}&  \cellcolor{blue!7}{arXiv 2025} &\cellcolor{blue!7}{8.03}&  \cellcolor{blue!7}{32.1}&  \cellcolor{blue!7}{59.4}&  \cellcolor{blue!7}{\textbf{62.8}}&  \cellcolor{blue!7}{\textbf{45.4}}&  \cellcolor{blue!7}{\textbf{77.1}}&  \cellcolor{blue!7}{59.4}&\\
\cellcolor{blue!15}{{Ming-Lite-Omni (ICL)}} \cite{ai2025ming}&  \cellcolor{blue!15}{arXiv 2025} &\cellcolor{blue!15}{8.15}&  \cellcolor{blue!15}{25.0}&  \cellcolor{blue!15}{41.9}&  \cellcolor{blue!15}{53.0}&  \cellcolor{blue!15}{38.4}&  \cellcolor{blue!15}{82.9 \textcolor{green}{$\uparrow$}}&  \cellcolor{blue!15}{41.9}&\\
   \midrule

   \cellcolor{blue!7}Video-LLaVA \cite{lin2024video}&  \cellcolor{blue!7}EMNLP 2024 &\cellcolor{blue!7}\textbf{3.18}& \cellcolor{blue!7}13.5&  \cellcolor{blue!7}\textbf{77.5}&  \cellcolor{blue!7}50.7&  \cellcolor{blue!7}14.3&  \cellcolor{blue!7}15.3&  \cellcolor{blue!7}\textbf{77.5}& \multirow{7}{*}{EgoPER}\\
   \cellcolor{blue!7}Video-LLaMA2 \cite{cheng2024videollama}&  \cellcolor{blue!7}arXiv 2024 &\cellcolor{blue!7}4.71 & \cellcolor{blue!7}13.6&  \cellcolor{blue!7}75.2&  \cellcolor{blue!7}51.6&  \cellcolor{blue!7}15.8&  \cellcolor{blue!7}18.9&  \cellcolor{blue!7}75.2& \\
   \cellcolor{blue!7}{Vinci} \cite{huang2025vinci}&  \cellcolor{blue!7}{IMWUT 2025} &\cellcolor{blue!7}{4.89}&  \cellcolor{blue!7}{12.0}&  \cellcolor{blue!7}{49.3}&  \cellcolor{blue!7}{47.4}&  \cellcolor{blue!7}{19.3}&  \cellcolor{blue!7}{\textbf{49.2}}&  \cellcolor{blue!7}{49.3}&\\
   \cellcolor{blue!7}LLaVA-OneVision \cite{lillava}&  \cellcolor{blue!7}TMLR 2025 &\cellcolor{blue!7}7.16& \cellcolor{blue!7}14.0&  \cellcolor{blue!7}60.1&  \cellcolor{blue!7}\textbf{54.5}&  \cellcolor{blue!7}21.3&  \cellcolor{blue!7}43.7&  \cellcolor{blue!7}60.1& \\
   \cellcolor{blue!7}{EgoGPT} \cite{yang2025egolife}&  \cellcolor{blue!7}{CVPR 2025} &\cellcolor{blue!7}{7.23}&  \cellcolor{blue!7}{\textbf{14.4}}&  \cellcolor{blue!7}{69.6}&  \cellcolor{blue!7}{51.3}&  \cellcolor{blue!7}{19.4}&  \cellcolor{blue!7}{29.8}&  \cellcolor{blue!7}{69.6}&\\
   \cellcolor{blue!7}{Qwen2.5-Omni-7B} \cite{xu2025qwen2}&  \cellcolor{blue!7}{arXiv 2025} &\cellcolor{blue!7}{9.61}&  \cellcolor{blue!7}{88.2}&  \cellcolor{blue!7}{21.8}&  \cellcolor{blue!7}{47.1}&  \cellcolor{blue!7}{29.0}&  \cellcolor{blue!7}{17.3}&  \cellcolor{blue!7}{21.8}&\\
   \cellcolor{blue!7}{Ming-Lite-Omni} \cite{ai2025ming}&  \cellcolor{blue!7}{arXiv 2025} &\cellcolor{blue!7}{13.2}&  \cellcolor{blue!7}{91.7}&  \cellcolor{blue!7}{18.6}&  \cellcolor{blue!7}{43.1}&  \cellcolor{blue!7}{\textbf{22.3}}&  \cellcolor{blue!7}{12.7}&  \cellcolor{blue!7}{18.6}&\\
   \bottomrule
   \end{tabular}
   \label{tab10}%
\end{table}

\subsection{Datasets and Implementation Details}

To more rigorously evaluate agents’ and AI assistants’ comprehension of egocentric procedural tasks, we conducted supplementary experiments on egocentric procedural error detection and egocentric procedural learning tasks. Four video question-answering models and two egocentric AI assistants were compared across four datasets to evaluate their proficiency in understanding procedural tasks.

\textit{Datasets.} To ensure broad scenario coverage, three egocentric datasets named CaptainCook4D \cite{peddi2024captaincook4d}, EgoPER \cite{lee2024error}, and EgoOops \cite{haneji2025egooops} were employed for egocentric procedural error detection tasks. CaptainCook4D and EgoPER address diverse kitchen activities, while EgoOops targets niche contexts such as chemical experiments and block-building. As these datasets lack predefined test sets, 20\% of CaptainCook4D and 30\% of EgoOops samples were randomly designated as test sets. For the procedural learning task, EgoProceL \cite{bansal2022my} was used, with PC assembly and disassembly tasks selected to assess the AI assistant’s competence in extracting critical procedural steps.

\textit{Implementation Details.} Both video question answering models and AI assistants are evaluated using question answering tasks. For egocentric procedural error detection, each model receives 4 to 5 randomly selected questions, with prompts tailored to highlight the error detection objective, thereby maximizing detection performance without fine-tuning. Following the protocol described in \cite{huang2025modeling}, keywords from model outputs are parsed to generate confidence scores, which are then used to calculate Precision, EDA, AUC, F1, Recall, and Accuracy. We also measure the average inference time (including sampling) for the first 50 samples. This comprehensive metric set ensures a robust assessment of error detection. Consistent prompt formats and confidence rules are applied across EgoPER and EgoOops, with minor adjustments for CaptainCook4D to accommodate model-specific attributes. To better show how different prompt strategies and context length affect results, we use the EgoOops dataset and test in‑context learning prompts that include definitions of possible error types and brief examples. For egocentric procedural learning, seven models were systematically evaluated on the PC assembly and PC disassembly tasks within EgoproceL, adhering to the experimental protocol established by \cite{mahmood2025procedure}. Performance was quantitatively assessed using the F1 score and Intersection over Union (IoU) metrics, and the results were benchmarked against existing procedural learning methodologies. To address relevant task requirements, tailored prompts were introduced. Given that the models lacked fine-tuning for key step learning and acknowledging the inherent challenge in uniformly evaluating semantically similar key step sequences, the egocentric procedural learning task was appropriately simplified. To mitigate excessive deviation between model predictions and expected outputs, a predefined set of candidate key steps was embedded in the prompts, enabling the models to select the sequence of steps that most closely matched each video segment. In accordance with the specifications of \cite{mahmood2025procedure}, and to enhance the objectivity of evaluating procedural learning abilities, the Hungarian matching algorithm was employed to analyze the result. In both experiments, we uniformly sample 16 frames from each step segment as input.

\subsection{Experimental Results}

The performance of egocentric procedural error detection models is evaluated using the metrics EDA, Accuracy, and Precision. In this study, EDA and Accuracy refer to the same concept, indicating the proportion of instances in which the model correctly classifies both normal and erroneous segments. This reflects the model’s overall ability to perform accurate classification. Precision measures the proportion of actual errors among all segments identified as erroneous by the model and serves as a key indicator of its effectiveness in detecting errors. Additional metric definitions are provided in previous sections.

Among the evaluated models, EgoGPT and Vinci are trained explicitly on egocentric datasets. Experimental results in Table~\ref{tab10} demonstrate that their classification performance on the CaptainCook4D dataset approaches random guessing, indicating limited detection capability. Other models, except Just Ask, exhibit only marginally better results. In terms of precision, these assistants significantly underperform specialized error detection models like AMNAR, reflecting a limited understanding of error identification and limiting their suitability for direct application in error detection tasks. Further analysis reveals a consistent bias toward classifying observed segments as correct, with prompt adjustments yielding slight improvement. This deficiency likely stems from insufficient exposure to error-containing data during training, impairing the models’ understanding and detection of procedural mistakes. Moreover, unlike standard anomaly detection, procedural error detection requires a comprehensive understanding of video content, as contextually correct actions may still constitute errors.

Experimental results on EgoOops and EgoPER show that only EgoGPT, trained on egocentric datasets, effectively understands errors, whereas other models yield low precision scores. Multimodal LLMs with strong reasoning capabilities, such as Ming-Lite-Omni, achieve high AUC, F1, and Recall scores, indicating strong overall detection performance. However, under identical token length and prompt settings, it tends to generate lengthy analytical responses. This hampers the accurate calculation of confidence scores. For the EDA metric, some approaches surpass 70, but this does not imply stronger classification ability than EgoPED and AMNAR. Instead, models generally classify the majority of segments as correct, where performance relies heavily on the dataset’s ratio of positive samples. Thus, a higher proportion of correct segments yields higher EDA scores, but this does not mean these models outperform EgoPED or AMNAR in classification ability. Under the in-context learning strategy, added context and examples in the prompt play a clear role, providing task-specific structure, clear error definitions, and simple examples of each error type, leading to substantial improvements in multiple metrics for the selected models. This demonstrates that prompt-based strategies can enhance models’ capability for procedural error detection to some extent, although there remains a performance gap compared with EgoPED and AMNAR, which in turn indicates a direction for future improvement.

In summary, through carefully designed experiments evaluating recent VQA models, we conclude that current models cannot be directly applied to egocentric procedural error detection tasks. Their understanding of errors is limited, indicating their overall video comprehension requires further improvement.

In the experimental evaluation of egocentric procedural learning shown in Table~\ref{tab11}, F1 and IoU metrics are utilized to assess the model’s capability. F1 score evaluates step recognition accuracy, while the IoU quantifies the temporal overlap between predicted steps and ground-truth segments.

Analysis of both PC assembly and disassembly tasks indicates that providing models with procedural knowledge does not yield satisfactory step recognition performance. However, in the context of PC disassembly and known procedural steps, Video-LLaMA2 outperforms two existing methods in step recognition accuracy, but still falls short of the state-of-the-art methods.

Regarding the IoU metric, LLaVA-OneVision demonstrates improved temporal localization accuracy in the PC assembly task, surpassing one method but still requiring enhancement. Video-LLaMA2 also attains strong IoU results in PC disassembly. Although EgoGPT and Vinci are pre-trained on egocentric datasets, they do not show marked advantages. As they lack task-specific fine-tuning for egocentric procedural error detection or learning, definitive comparisons with other models are unwarranted. Further rigorous experiments are needed to assess their capabilities thoroughly. Finally, in the error detection task, Video-LLaVA achieves the shortest average inference time, while in procedural learning, Vinci is the fastest. Still, both operate on the order of seconds, which may be insufficient for providing real-time and effective support.

In conclusion, current VQA models are not directly applicable to egocentric procedural learning tasks. Even with simplified step recognition settings, their overall performance remains unsatisfactory, despite some models showing potential in step identification.

\begin{table}
\caption{Experimental results in egocentric procedural learning, the part highlighted in purple is the result of the model we selected, bold text represents the best performance on this metric in each part, and Time is the average inference time.}
    \centering
    \begin{tabular}{ccccc|c|ccc|c|}
    \toprule
    \multicolumn{1}{c|}{ \cellcolor[HTML]{EFEFEF}Method}&   \multicolumn{1}{c|}{ \cellcolor[HTML]{EFEFEF}Years} & \multicolumn{1}{c|}{ \cellcolor[HTML]{EFEFEF}Time}&   \multicolumn{1}{c|}{ \cellcolor[HTML]{EFEFEF}F1}&   \multicolumn{1}{c|}{ \cellcolor[HTML]{EFEFEF}IoU}&   \multicolumn{1}{c|}{ \cellcolor[HTML]{EFEFEF}Task} &\multicolumn{1}{c|}{ \cellcolor[HTML]{EFEFEF}Time}&\multicolumn{1}{c|}{ \cellcolor[HTML]{EFEFEF}F1}&   \multicolumn{1}{c|}{ \cellcolor[HTML]{EFEFEF}IoU}&   \multicolumn{1}{c|}{ \cellcolor[HTML]{EFEFEF}Task}\\
    \midrule
         CnC \cite{bansal2022my}&  ECCV 2022 &-&  25.1&  12.8& \multirow{10}{*}{PC Assembly} &-&  27.0&  14.8& \multirow{10}{*}{PC Disassembly}\\
         OPEL \cite{chowdhury2024opel}&  NeurIPS 2024 &-&  33.7&  17.9&  &-&32.2&  16.9& \\
         RGWOT \cite{mahmood2025procedure}&  arXiv 2025 &-&  \textbf{43.6}&  \textbf{28.0}&  &-&\textbf{45.9}&  \textbf{30.1}& \\

            \cellcolor{blue!7}Video-LLaVA \cite{lin2024video}&  \cellcolor{blue!7}EMNLP 2024 &\cellcolor{blue!7}5.34& \cellcolor{blue!7}\textbf{22.9}&  \cellcolor{blue!7}12.4&  &\cellcolor{blue!7}6.03& \cellcolor{blue!7}10.0&  \cellcolor{blue!7}4.9&\\
            \cellcolor{blue!7}Video-LLaMA2 \cite{cheng2024videollama}&  \cellcolor{blue!7}arXiv 2024 &\cellcolor{blue!7}4.85& \cellcolor{blue!7}12.0&  \cellcolor{blue!7}6.9&  &\cellcolor{blue!7}5.44& \cellcolor{blue!7}\textbf{35.8}&  \cellcolor{blue!7}\textbf{21.6}&\\
                \cellcolor{blue!7}{Vinci} \cite{huang2025vinci}&  \cellcolor{blue!7}{IMWUT 2025} &\cellcolor{blue!7}{\textbf{4.75}}&  \cellcolor{blue!7}{14.1}&  \cellcolor{blue!7}{7.5}&  &\cellcolor{blue!7}{\textbf{4.83}}&  \cellcolor{blue!7}{27.2}&  \cellcolor{blue!7}{14.7}&\\
            \cellcolor{blue!7}LLaVA-OneVision \cite{lillava}&  \cellcolor{blue!7}TMLR 2025 &\cellcolor{blue!7}7.18& \cellcolor{blue!7}22.2&  \cellcolor{blue!7}\textbf{13.0}&  &\cellcolor{blue!7}7.36& \cellcolor{blue!7}16.8&  \cellcolor{blue!7}8.8&\\
            \cellcolor{blue!7}{EgoGPT} \cite{yang2025egolife}&  \cellcolor{blue!7}{CVPR 2025} &\cellcolor{blue!7}{6.86}&  \cellcolor{blue!7}{6.6}&  \cellcolor{blue!7}{3.7}&  &\cellcolor{blue!7}{7.57}&  \cellcolor{blue!7}{8.9}&  \cellcolor{blue!7}{4.3}&\\
            \cellcolor{blue!7}{Qwen2.5-Omni-7B} \cite{xu2025qwen2}&  \cellcolor{blue!7}{arXiv 2025} & \cellcolor{blue!7}{15.0}&  \cellcolor{blue!7}{9.6}&  \cellcolor{blue!7}{5.6}&   & \cellcolor{blue!7}{15.8}&\cellcolor{blue!7}{1.5}&  \cellcolor{blue!7}{0.8}&\\
 \cellcolor{blue!7}{Ming-Lite-Omni} \cite{ai2025ming}&  \cellcolor{blue!7}{arXiv 2025} &\cellcolor{blue!7}{17.4}&  \cellcolor{blue!7}{9.3}&  \cellcolor{blue!7}{5.9}&   &\cellcolor{blue!7}{16.6}&\cellcolor{blue!7}{7.6}&  \cellcolor{blue!7}{4.5}&\\
            \bottomrule
    \end{tabular}
    
    \label{tab11}%
\end{table}

\section{Challenges and Outlook}

\subsection{Challenges}

Our extensive experiments reveal that current vision-language models, including egocentric AI assistants, are insufficient for supporting procedural tasks. This highlights the necessity for further innovation to enable effective AI facilitation of such processes. Although notable progress has been made, egocentric vision tasks face challenges as they transition from controlled settings to real-world applications.

\textit{\textbf{System-level bottlenecks, latency, and deployment constraints.}} Although we have analyzed real-time and streaming video understanding to reduce latency, major challenges remain in deploying models on wearable devices \cite{lane2016deepx}. Many VLMs are too large for current hardware and, even when deployable, still suffer from latency, overheating, and related issues. Larger models usually deliver better reasoning and performance, and added functionality increases resource demands. Balancing model size with deployment constraints is thus a central challenge for realizing practical, real-world AI assistants.

\textit{\textbf{Data scarcity and bias.}} Existing egocentric video datasets are scarce, lack diversity, and provide limited annotation detail, constraining both error detection and procedural learning. While many procedural learning datasets are captured from third-person perspectives, they are ill-suited for egocentric tasks. Most available egocentric error detection datasets also cover narrow ranges of activities, failing to support comprehensive AI training for a wide range of real-world tasks. Additionally, cultural and individual biases present in current datasets further hinder robust generalization. Human annotation dependence limits the practicality of real-world applications, constrains real-time detection, and narrows application domains. Recent egocentric generation methods, such as diffusion-based models like LEGO \cite{lai2024lego}, can be used to augment datasets and mitigate data scarcity, annotation gaps, and inherent biases.

\textit{\textbf{Limited understanding of long-term procedural tasks.}} Current models are struggling to capture logical and temporal dependencies in procedural videos, failing to distinguish true procedural errors or provide actionable explanations. They encounter difficulties in computational efficiency, semantic understanding, and the effective integration of multimodal signals, which hampers comprehensive video interpretation. Existing approaches, such as error detection, rely on a foundational model comprehension of procedural steps. Future research should prioritize adapting advanced long-video understanding frameworks \cite{goletto2024amego,plizzari2025spatial} to egocentric contexts and enhancing multimodal fusion \cite{xu2025qwen2,ai2025ming,xu2025qwen3omnitechnicalreport,xie2024miniomni2opensourcegpt4ovision,meituanlongcatteam2025longcatflashomnitechnicalreport} for deeper procedural task analysis and improved model support.

\subsection{Egocentric Procedural Reasoning, Planning, and Action Anticipation}

Despite existing limitations, research on egocentric vision and AI assistants is advancing rapidly, with expanding domains and growing scholarly interest.

\textit{\textbf{Egocentric procedural reasoning}} \cite{vinod2025egovlm,lee2025towards,peirone2025hier}.  It is an emerging area aiming for deep causal, temporal, and logical inference over egocentric video content. Recent research prioritizes understanding action relationships, accurate episodic memory retrieval, social interaction, and hand-object analysis \cite{su2024care,su2025annexe,deng2025egocentric,wang2019object}, moving beyond simple associative learning. Given the preceding results, current models and assistants show limited competence in procedural tasks. Improving their reasoning through in-context learning, chain-of-thought prompting, or multi-agent systems with deliberate reasoning and tool-use capabilities may effectively address these procedural challenges.

\textit{\textbf{Egocentric procedural planning and action anticipation.}} Recent work \cite{zhao2023antgpt,fang2024egocentric,kulkarni2025egovita,liu2023egocentric,rhinehart2017first,rodin2024action} increasingly strengthens models’ procedural understanding by recognizing past and ongoing actions to infer and predict future actions and action sequences. For EgoProceAssist, this paradigm is not only a key foundation but also a promising route to tackling procedure-centric tasks. Prior studies such as PREGO \cite{flaborea2024prego}, TI-PREGO \cite{plini2025ti}, and AMNAR \cite{huang2025modeling} detect errors by predicting the next plausible action at each timestep, while GC \cite{mazzamuto2025gazing} predicts future gaze from eye-movement trajectories and flags errors via gaze deviations, achieving accurate error detection with real-time responsiveness without processing the full video. These results show that anticipation provides a novel mechanism for error detection and enables inferring user intent from signals such as gaze, which is crucial for building practical assistants. In procedural learning and QA, anticipation supports online updating of step sequences, offering new ways to monitor task progress and verify results. By anticipating user intent, the assistant can proactively deliver support, plan feasible action paths, and provide targeted guidance. Consequently, egocentric procedural planning and action anticipation emerge as a central direction that warrants sustained and systematic investigation.

\subsection{Edge AI, Model Distillation, and Cloud-Edge Architectures}

Recent studies employ lightweight backbone networks with feature compression and streaming video processing to support deployment on smart glasses and to exploit fog and cloud services. Future work can prioritize edge AI and on-device inference. Lightweight architectures and hardware-aware optimizations are key to deploying AI assistants on resource-constrained devices. Model compression techniques such as pruning \cite{zhu2017prune}, quantization \cite{han2015deep}, and knowledge distillation \cite{hong2022analysis}, have been applied to VLMs to reduce size and latency with minimal accuracy loss, including distilling large VLMs into smaller, task-specific models. Cloud–edge collaborative architectures \cite{yu2025cloud} are also promising. Hybrid systems run lightweight models on-device for latency-critical tasks (e.g., error detection) and offload complex reasoning to the cloud, balancing latency and computational load.

Integrating techniques from these domains can significantly enhance models' comprehension of procedural tasks, leading to substantial performance improvements and enabling AI assistants to better support our daily lives.

\section{Conclusion}

This work conceptualizes an egocentric procedural AI assistant (EgoProceAssist), structuring the field into three core areas: egocentric procedural error detection, egocentric procedural learning, and egocentric procedural question answering, along with two enabling dimensions: real-time and streaming video understanding, and proactive interaction in procedural contexts. Existing literature is systematically reviewed and categorized. 
A novel taxonomy is introduced that provides a systematic classification and synthesis of existing methodological approaches. Subsequently, a thorough examination of the prevalent datasets and evaluation metrics used to assess the three principal tasks is presented. In addition, a new experimental paradigm is devised to rigorously evaluate the performance of generative AI assistants in the domains of egocentric procedural error detection and procedural learning. Empirical findings substantiate that extant AI assistants are presently insufficient for the direct support of procedural tasks. Lastly, the study delineates the persistent challenges and inherent limitations in egocentric vision research, with a particular emphasis on the construction of intelligent agents, and elucidates prospective avenues that may significantly influence subsequent advances in the field. We hope this work informs and inspires further research within the community.

\section{Acknowledgment}

The research work described in this paper was conducted in the JC STEM Lab of Machine Learning and Computer Vision funded by The Hong Kong Jockey Club Charities Trust. This research received partially support from the Global STEM Professorship Scheme from the Hong Kong Special Administrative Region.

\section{Data Availability Statement}

The datasets used for the supplementary experiment are all publicly available; however, to obtain EgoPER, you must email the author. The datasets we show in Table~\ref{tab2}, ~\ref{tab4}, ~\ref{tab6}, ~\ref{tab8}, and ~\ref{tab9} are all publicly available. The experiment data shown in Table~\ref{tab1}, ~\ref{tab3}, ~\ref{tab5}, and ~\ref{tab7} are all publicly available. To make it easier to use, we have summarized them in \url{https://github.com/z1oong/Building-Egocentric-Procedural-AI-Assistant}.

\bibliography{reference}
%\bibliographystyle{plain}
%\bibliography{reference}

\end{document}